\documentclass{article}

\usepackage[preprint]{neurips_2026}

\usepackage[utf8]{inputenc} 
\usepackage[T1]{fontenc}    
\usepackage{hyperref}       
\usepackage{url}            
\usepackage{booktabs}       
\usepackage{amsfonts}       
\usepackage{nicefrac}       
\usepackage{microtype}      
\usepackage{xcolor}         

\usepackage{graphicx}
\usepackage{enumitem}
\usepackage{multirow}
\newcommand{\dataname}{\textsc{DocPair}}
\newcommand{\benchname}{\textsc{DocPairBench}}
\newcommand{\modelname}{\textsc{DocReward}}
\newcommand{\benchpaircount}{1443}
\usepackage{amsmath}

\usepackage{amsmath,amsfonts,bm}

\def\eqref#1{equation~\ref{#1}}

\def\1{\bm{1}}

\DeclareMathAlphabet{\mathsfit}{\encodingdefault}{\sfdefault}{m}{sl}
\SetMathAlphabet{\mathsfit}{bold}{\encodingdefault}{\sfdefault}{bx}{n}

\usepackage{subcaption}

\usepackage[most]{tcolorbox} 
\definecolor{lightgreen}{RGB}{145, 204, 117}
\tcbuselibrary{listings,skins,breakable} 
\newtcblisting{myverbatimbox}[1][Prompt]{
  colback=lightgray!20, 
  colframe=lightgreen, 
  fontupper=\ttfamily, 
  title=#1, 
  breakable,
  enhanced,
  listing only, 
  boxsep=5pt,
  left=0pt,
  right=0pt,
  top=0pt,
  bottom=0pt,
}
\usepackage{wrapfig}

\title{\modelname{}: A Document Reward Model for Structuring and Stylizing}

\author{Junpeng Liu\textsuperscript{1}\thanks{~Equal contribution. Work done during internship at Microsoft Research.}\;
Yuzhong Zhao\textsuperscript{2}\footnotemark[1]\;
\textbf{Bowen Cao}\textsuperscript{1}\;
\textbf{Jiayu Ding}\textsuperscript{3}\;
\textbf{Yilin Jia}\textsuperscript{4}\;
\textbf{Tengchao Lv}\textsuperscript{5} \\
\textbf{Yupan Huang}\textsuperscript{5}\; 
\textbf{Wenshan Wu}\textsuperscript{5}\; 
\textbf{Shaohan Huang}\textsuperscript{5}\; 
\textbf{Nan Yang}\textsuperscript{5}\; 
\textbf{Li Dong}\textsuperscript{5}\; 
\textbf{Lei Cui}\textsuperscript{5}\; 
\textbf{Tao Ge}\textsuperscript{5} \\
\textbf{Xun Wang}\textsuperscript{5}\; 
\textbf{Huitian Jiao}\textsuperscript{5}\;
\textbf{Sun Mao}\textsuperscript{5}\;
\textbf{FNU Kartik}\textsuperscript{5}\;
\textbf{Si-Qing Chen}\textsuperscript{5}\;
\textbf{Wai Lam}\textsuperscript{1}\;
\textbf{Furu Wei}\textsuperscript{5}  \\
\textsuperscript{1}CUHK
\quad
\textsuperscript{2}UCAS
\quad
\textsuperscript{3}XJTU
\quad
\textsuperscript{4}UMich
\quad
\textsuperscript{5}Microsoft \\
{\href{https://aka.ms/GeneralAI}{https://aka.ms/GeneralAI}}
}

\begin{document}

\maketitle

\begin{abstract}
Recent agentic workflows automate professional document generation but focus narrowly on textual quality, overlooking structural and stylistic professionalism, which is equally critical for readability. 
This gap stems mainly from a lack of effective reward models capable of guiding agents toward producing documents with high structural and stylistic professionalism.
We introduce \modelname{}, a \textit{\textbf{Doc}ument \textbf{Reward} Model} that evaluates documents based on their structure and style.
To achieve this, we propose a \textit{textual-quality-agnostic} framework that ensures assessments are not confounded by content quality, and construct \dataname{}, a dataset of 117K paired documents covering 32 domains and 267 types. Each pair shares identical content but differs in structural and stylistic professionalism.
\modelname{} is trained using the Bradley-Terry loss. 
On a manually annotated benchmark, \modelname{} outperforms GPT-5 by \textbf{14.6} percentage points in the same setting.
Reinforcement learning experiments further show that \modelname{} effectively guides agents toward generating documents with consistently higher structural and stylistic professionalism, highlighting its practical utility.
\end{abstract}

\section{Introduction}
\label{sec:introduction}
\begin{figure*}[h]
  \includegraphics[width=1.0\linewidth]{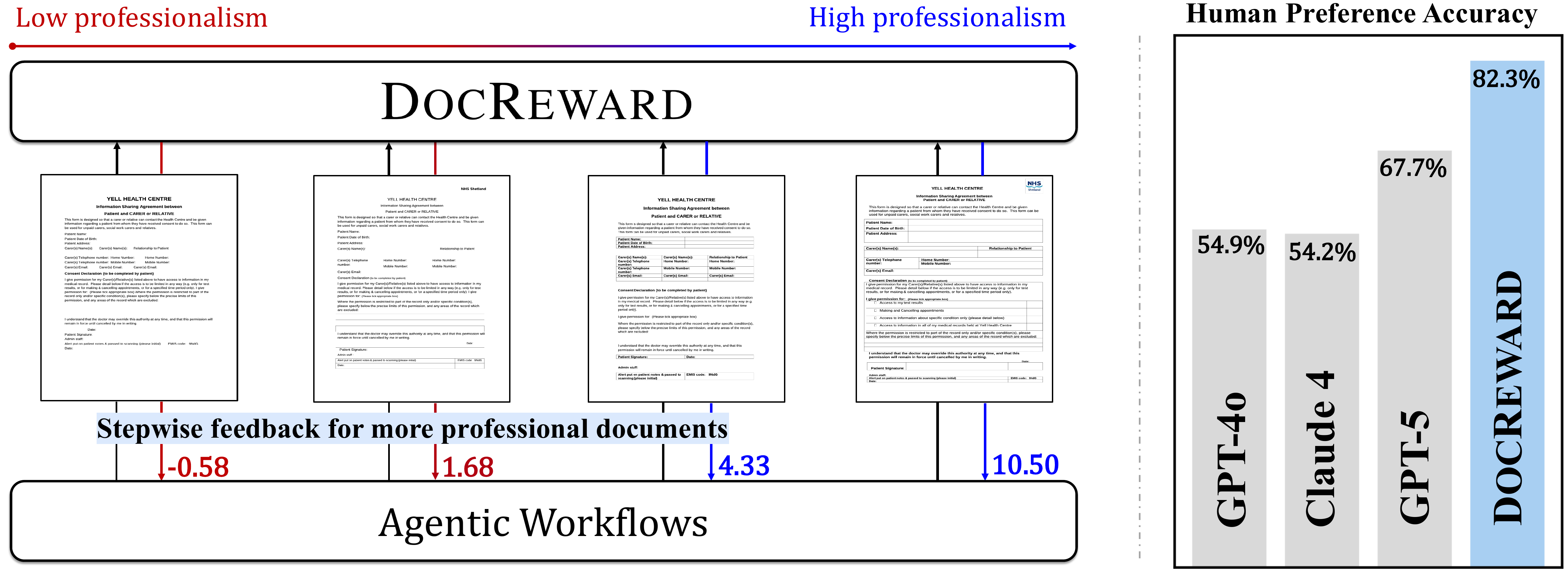}
     \caption{\modelname{} automatically assesses document professionalism according to the structure and style of documents, assisting existing agentic workflows for more professional document generation (left). It outperforms GPT-5 by 14.6 percentage points in human preference accuracy in the same setting (right).
     }
    \label{fig:1}
\end{figure*}


Recent advances in agentic workflows have automated various complex tasks, such as code generation~\citep{peng2023copilot,anthropic2025claudecode,hong2024metagpt}, image generation~\citep{Xiao_2025_CVPR}, visual understanding~\citep{zheng2025deepeyes,Marsili2025Visual_agentic_ai}, math reasoning~\citep{yan2025mathagent}, and travel planning~\citep{xie2024travelplanner}. A key focus of agentic workflows is the generation of professional documents, such as deep research~\citep{openai2025deepresearch,openmanus2025,qwen2025deepresearch} and technical documentation generation~\citep{dvivedi2024code_documentation}. However, existing research on professional document generation primarily focuses on improving textual quality, often neglecting the visual structure and style—both of which are crucial for document professionalism.
A well-organized structure facilitates seamless navigation for readers, while a consistent style makes the content more readable and engaging. Together, these aspects help convey information more clearly and effectively.
The neglect of structure and style mainly stems from the lack of effective reward models, which are capable of guiding agentic workflows toward producing documents with professional structure and style.



%
However, building a reward model capable of providing a robust evaluation of visual structure and style is non-trivial, as it requires both \textit{comprehensiveness} and \textit{textual-quality-agnosticism}. Specifically, comprehensiveness refers to the ability to evaluate documents across diverse types, qualities, structures, and styles, while textual-quality-agnosticism, in this context, means that the model does not evaluate the inherent quality of the textual content itself, but instead assesses how well the structure and style of a document stand out, given the fixed content.

To achieve this, we propose \modelname{}, a \textit{\textbf{Doc}ument \textbf{Reward} Model} specialized in assessing document professionalism in structure and style, as illustrated in~\autoref{fig:1}.
The model is trained under a textual-quality-agnostic framework. Specifically,
we construct \dataname{}, a multi-domain dataset of 117K document pairs across 32 domains and 267 types. Each pair comprises a high-professionalism sample and its low-professionalism counterpart; these documents share identical content but differ in structure and style.
The construction of \dataname{} involves three phases: 
1) \textit{Curating High-Quality Professional Documents.} We curate a collection of high-quality documents characterized by professional structure and style from diverse domains, such as government, education, and science.
2) \textit{Expanding Source Documents via Agents.} Next, we extract the textual content and employ multiple generation agents to re-produce documents that preserve the original textual content while exploring diverse structural and stylistic variations.
%
3) \textit{Ranking Documents.} The ranking labels are determined by a combination of human-verified heuristics and an oracle-based annotation method. This provides a scalable way to capture elements of structural and stylistic professionalism.


Based on the constructed dataset,  we train \modelname{} to take rendered document pages as input and output a score reflecting the document's professionalism in structure and style. 
The predicted scores of paired documents are optimized using the Bradley-Terry loss~\citep{bradley1952rank,ouyang2022training}, which penalizes violations of the annotated ranking order.

To demonstrate the superiority and utility of \modelname{}, we conduct both intrinsic and extrinsic evaluations. 
For the intrinsic evaluation, we establish a benchmark \benchname{} of \benchpaircount{} human-annotated pairs across multiple domains. Human annotators ranked each pair based on the professionalism of the documents' structure and style. Notably, as shown in~\autoref{fig:1} (right), \modelname{} outperforms GPT-4o~\citep{hurst2024gpt4o} and GPT-5~\citep{openai2025gpt5} by 27.4 and 14.6 percentage points, respectively, in accuracy on \benchname{}, demonstrating its superiority over existing baselines. 
For the extrinsic evaluation, we evaluate \modelname{} through two complementary experiments. 
1) \textit{Best-of-N}. \modelname{} is used as a re-ranking model for improving agentic workflow without changing the agent itself. Human evaluation reveals that \modelname{} as a reward model achieves a significantly higher win rate of 60.8\%, compared to GPT-5's 37.7\%. 
%
2) \textit{Reinforcement Learning}. We further demonstrate the utility of \modelname{} as the reward model for reinforcement learning of both open- and closed-source models. This integration improves the document generation performance of Qwen2.5-Coder~\citep{hui2024qwen25coder} and GPT-4o in terms of structure and style.
%

Our main contributions are as follows:

\begin{itemize}[leftmargin=*]
    \item We propose a \textit{textual-quality-agnostic} framework for document reward modeling. By decoupling structure and style from textual content, it enables the reward model \modelname{} to accurately capture complex structural and stylistic elements.

    \item \dataname{}: A large-scale multi-domain dataset comprising 117K document pairs across 32 domains and 267 document types, designed to equip \modelname{} with comprehensiveness and textual-quality-agnosticism. 

    \item Comprehensive experiments demonstrate that \modelname{} not only outperforms GPT-5 in assessing structure and style in the same setting, but also serves as an effective reward model in reinforcement learning for agentic workflows.
\end{itemize}




\section{Textual-Quality-Agnostic Framework}
\label{sec:task}


Document professionalism is characterized by its textual content, structure, and style. Although large language models excel at evaluating textual quality, they are limited in assessing structure and style. 
To bridge this gap, we propose a \textit{textual-quality-agnostic} framework that trains reward models to evaluate professionalism independently of textual quality.




In this section, we formulate the framework and define its objectives.
Let $\{D_i\}_{i=1}^{N}$ denote a set of $N$ documents, where each document $D_i$ consists of textual content $D_{\mathrm{text}, i}$ and rendered images $D_{\mathrm{img}, i}$. The document reward model $\mathcal{R}_\theta$ assigns scores to documents that share the same textual content, such that these scores reflect their structural and stylistic quality. This process is formalized as follows:

\vspace{-3mm}
{
\small
\begin{align}
\max_{\theta} \; \mathrm{Sim} \big( \pi^*, \mathrm{Argsort}( & \mathcal{R}_\theta(D_{img,1}), \dots, \mathcal{R}_\theta(D_{img,N})) \big)\notag \\ 
\text{s.t. } D_{\mathrm{text}, i} = D_{\mathrm{text}, j}, &\quad \forall i, j,
\end{align}
}
where ``$\mathrm{Sim}$'' is a predefined similarity function that measures the agreement between the true and predicted orders. ``$\mathrm{Argsort}$'' returns the indices of documents sorted by their predicted scores. 
$\pi^*$ denotes the true indices reflecting the relative ranking of the documents in terms of structure and style.
Essentially, the term ``$D_{\mathrm{text}, i} = D_{\mathrm{text}, j}$’’ ensures that the model evaluates professionalism in a \textit{textual-quality-agnostic} manner, as it processes identical textual content from paired documents. Importantly, this framework is content-quality-independent rather than content-agnostic: the model still reads and understands the textual content to judge whether the structure and style are contextually appropriate, but its judgment is not influenced by how well the text itself is written.

In this paper, document professionalism in structure and style is defined as follows:

\textit{1) Structure:} Proper use of white space, appropriate margins, clear section breaks, well-structured text alignment, adequate paragraph spacing, proper indentation, inclusion of page headers and footers, and logical, coherent organization of content.

\textit{2) Style:} Appropriate font choices (type, size, color, readability), clear heading styles, effective use of emphasis (bold, italics), bullet points, numbering, and consistent formatting.


By optimizing $\mathcal{R}_\theta$ based on these factors, we obtain a reward model capable of assessing structural and stylistic professionalism in a comprehensive and textual-quality-agnostic way.

\section{DocReward}

Based on the proposed \textit{textual-quality-agnostic} framework, we train \modelname{}, a reward model specializing in assessing the structural and stylistic professionalism of documents. \modelname{} is trained on \dataname{}, a diverse dataset of 117K document pairs (Section~\ref{sec:data_construction}), and is optimized with a preference-based objective for structural and stylistic assessment (Section~\ref{sec:model}). The following sections detail the data construction pipeline and model design.

\begin{figure*}[t]
	\includegraphics[width=1.0\linewidth]{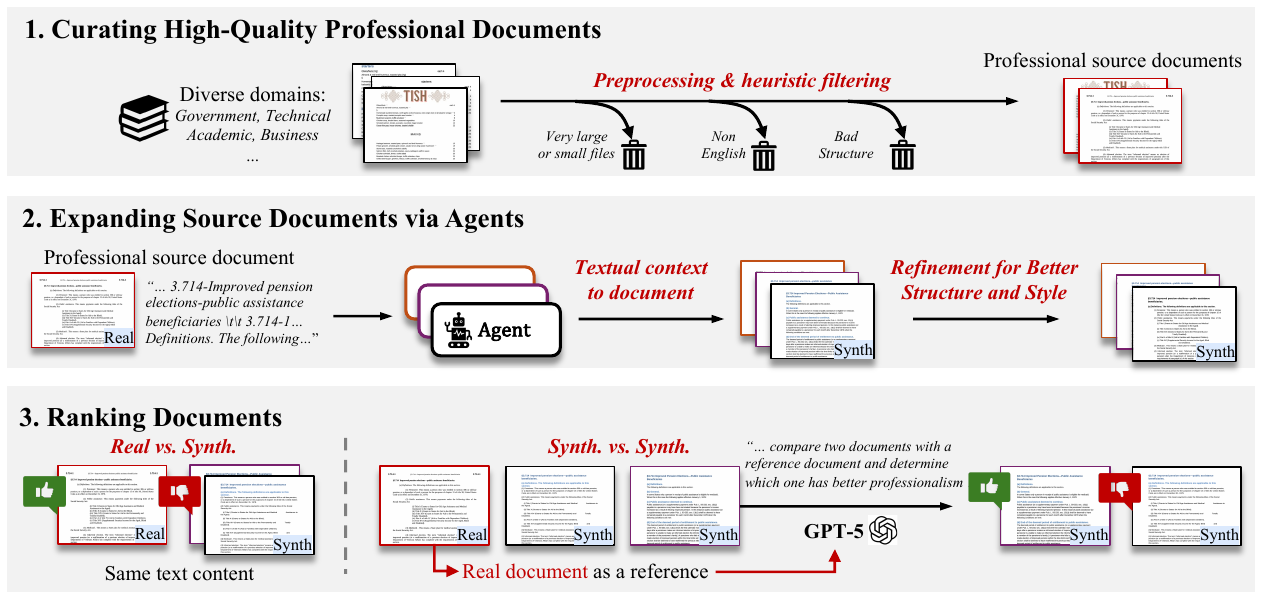}
     \caption{The data construction pipeline for \modelname{}.}
    \label{fig:data_construction}
\end{figure*}

\subsection{Data Construction}
\label{sec:data_construction}
As shown in~\autoref{fig:data_construction}, we first collect a set of high-quality real-world source documents.
The source documents are then expanded by multiple generation agents, and the resulting documents are grouped by shared textual content. 
Finally, each group of documents is annotated with a ranking $\pi^*$ in terms of structure and style quality.
%
The construction procedure is detailed step by step below:



\noindent\textbf{Curating High-Quality Professional Documents.} As illustrated in~\autoref{fig:data_construction} (top), we first curate a corpus of human-authored Microsoft Word documents that spans both highly formal institutional writing and everyday professional communication. We draw on two complementary sources:

\textit{1) Government corpora:} GovDocs1~\citep{garfinkel2009govdocs} and NapierOne~\citep{davies2022napierone} are authoritative document collections sourced from government and public institutions, covering reports, forms, guidelines, and other professional materials with consistent structure and style.

\textit{2) Web document corpus:} CommonCrawl, covering business, education, law, healthcare, \emph{etc.}\footnote{\url{https://commoncrawl.org/}} 

To ensure high quality, we apply a preprocessing and filtering pipeline before data construction. First, all files are converted to DOCX format to enable programmatic access and modification via \texttt{python-docx}.\footnote{\url{https://python-docx.readthedocs.io/}} Next, we discard extreme or malformed cases (exceeding 20 pages, files larger than 1\,MB dominated by images, and files smaller than 10\,KB with trivial content). To efficiently reduce residual noise, we employ GPT-5 as a rigorous automated heuristic to flag poor structure/style on a $[0,10]$ scale; documents scoring above 8 are retained. A manual inspection of 200 randomly sampled retained documents confirms that this automated filter preserves high-quality professional samples. 

\begin{wrapfigure}{r}{0.4\textwidth}
        \centering
        \vspace{-4mm}
        \includegraphics[width=0.4\textwidth]{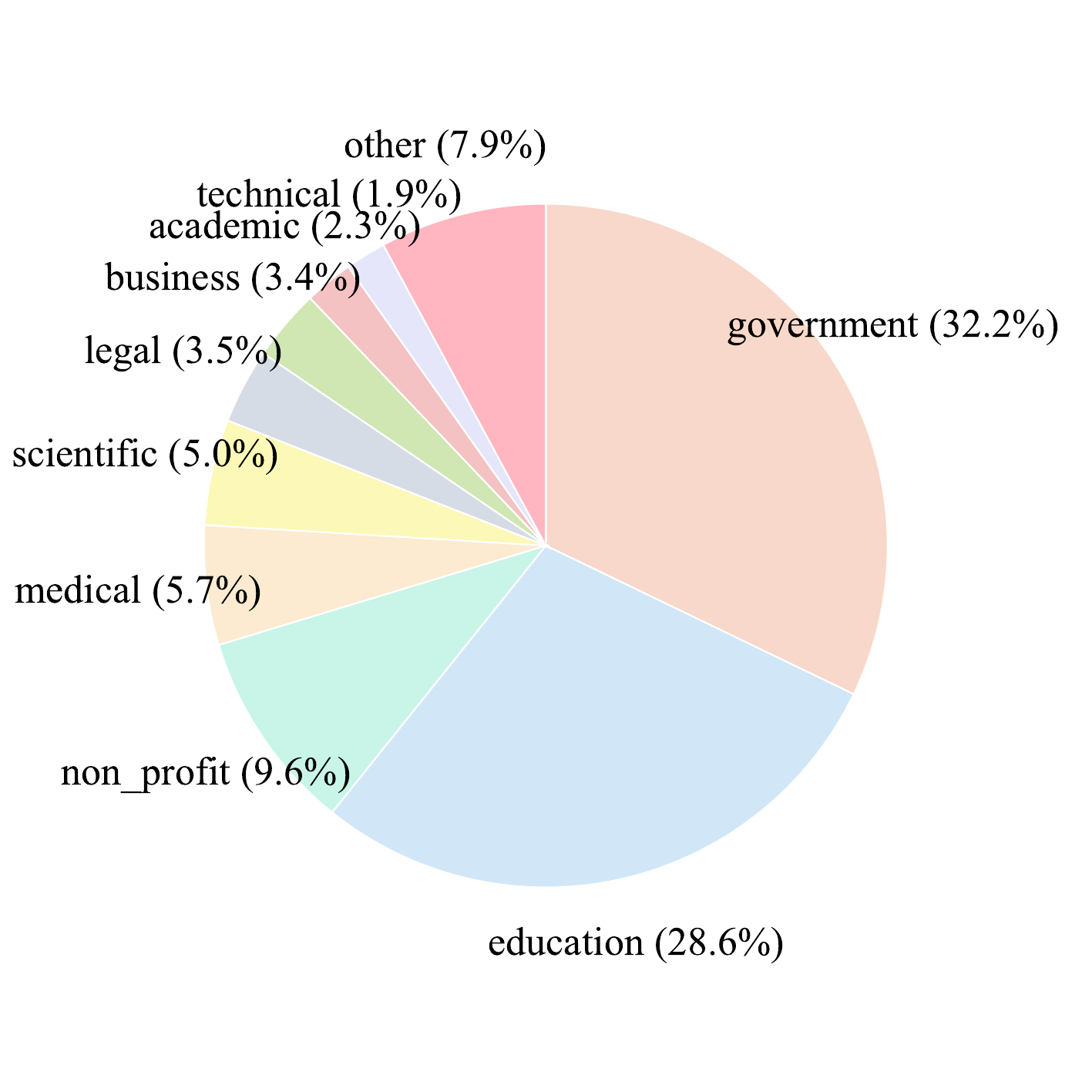}
        \vspace{-6mm}
        \caption{Top 10 domain distribution.}
        \label{fig:domain_dist}
        \vspace{-4mm}
    \end{wrapfigure}

The domain distributions of the filtered corpus are shown in~\autoref{fig:domain_dist}. The corpus spans 32 domains and 267 document types, demonstrating substantial breadth and diversity, and it provides a high-quality foundation for constructing subsequent document pairs. More data statistics are presented in Appendix~\ref{appendix:data_statistics}.

\noindent\textbf{Expanding Source Documents via Agents.} 
\label{sec:expand_documents_via_agent}
As shown in~\autoref{fig:data_construction}(middle), to obtain documents with the same textual content but different structure and style, we construct two types of agents to synthesize documents given the textual content (and rendered pages) of the source documents. To further increase the diversity of the synthesized documents, each agent can be empowered by different LLMs. The two agents are detailed as follows:

\textit{1) Textual Content to Document.} 
The textual content is first extracted from the source documents, discarding all formatting, styling, and layout information. Then, advanced models ($e.g.$, GPT-4o, OpenAI o1~\citep{openai2024o1system}, Claude Sonnet 4~\citep{anthropic2025claude4}, and GPT-5) are used to synthesize DOCX documents via \texttt{python-docx}. 

\textit{2) Refinement for Better Structure and Style.} 
To further improve the structure and style of synthesized documents, we refine them by comparing them with the original human-authored documents in terms of structure and style. The refinement process consists of two stages: 
a) Agents are provided with the \texttt{python-docx} code, rendered pages, and structured textual representation of the synthesized document, along with the rendered pages of the original human-authored document, to generate a refinement plan.  
b) Based on the plan, the agents modify the \texttt{python-docx} code to produce refined documents.  
    
The synthesized documents are grouped with their originals to facilitate subsequent processing. Please refer to Appendix~\ref{appendix:data_construction_detail} and~\ref{appendix:prompts} for details.

\noindent\textbf{Ranking Documents.}
As shown in~\autoref{fig:data_construction} (bottom), the collected documents within each group share identical textual content and are organized into pairs. In the ranking stage, we assess the relative professionalism in terms of structure and style for each pair, under the following two cases:

\textit{1) Real vs. Synth.} 
When comparing an original document with its agent-generated counterparts, the human-authored version is designated as the winner. This heuristic is grounded in a human inspection (on a randomly sampled set of 100 samples), which indicates that current state-of-the-art models (e.g., GPT-5, Claude Sonnet 4) produce documents with structural and stylistic quality inferior to high-quality human-authored documents curated by the rigorous filtering pipeline described earlier.

\textit{2) Synth. vs. Synth.} For pairs where both documents are agent-generated, manual annotation is not scalable. We therefore employ a closed-source model as a proxy for human judgment. To mitigate its intrinsic biases, we adopt an oracle setting: the model is provided with a document triplet $\{D_\mathrm{real}, D_\mathrm{synth1}, D_\mathrm{synth2}\}$, using the human-authored $D_\mathrm{real}$ as a reference anchor. This transforms the annotation from a subjective preference task into an objective similarity matching against the ground-truth document (see the prompt in Appendix~\ref{sec:triple_wise_prompt}). 
We validated this approach on 120 pairs, where it demonstrated 92.5\% alignment with human experts, confirming its effectiveness.

\begin{wraptable}{r}{0.45\textwidth}
\centering
\vspace{-1em}
\caption{Data statistics of \dataname{}.}
\small
{
\setlength{\tabcolsep}{3pt}
\begin{tabular}{ccccc}
\toprule
Domains & Types & Doc. & Avg. Page & Pairs \\
\midrule
32 & 267 & 69,137 & 3.2 & 117,108 \\
\bottomrule
\end{tabular}
}
\label{tab:data_statistics}
\vspace{-1em}
\end{wraptable}
Note that the above strategy is cost-effective and necessary for large-scale training data construction, while the test split, \benchname{}, is constructed through strict human annotation. The data statistics of \textsc{DocPair} are shown in~\autoref{tab:data_statistics}.

\subsection{Model Structure and Optimization}
\label{sec:model}
%
We use Qwen2.5-VL~\citep{bai2025qwen25vl} as the base model, which takes a multi-page document as input and outputs a scalar score via an added regression head (details provided in Appendix~\ref{appendix:model_impl_detail}).

Given a preference pair consisting of a winner $D_{\mathrm{img}}^w$ and a loser $D_{\mathrm{img}}^l$, the reward model $\mathcal{R}_\theta$ assigns scores to both documents and is optimized using the Bradley-Terry (BT) loss:
\begin{equation}
   \min_\theta -\log \sigma\big(\mathcal{R}_\theta(D_{\mathrm{img}}^w) - \mathcal{R}_\theta(D_{\mathrm{img}}^l)\big),
\end{equation}
where $\sigma(x)=\frac{1}{1+e^{-x}}$. This objective encourages higher scores for preferred documents.



\section{Experiments}
\begin{figure*}[]
        \includegraphics[width=1.0\linewidth]{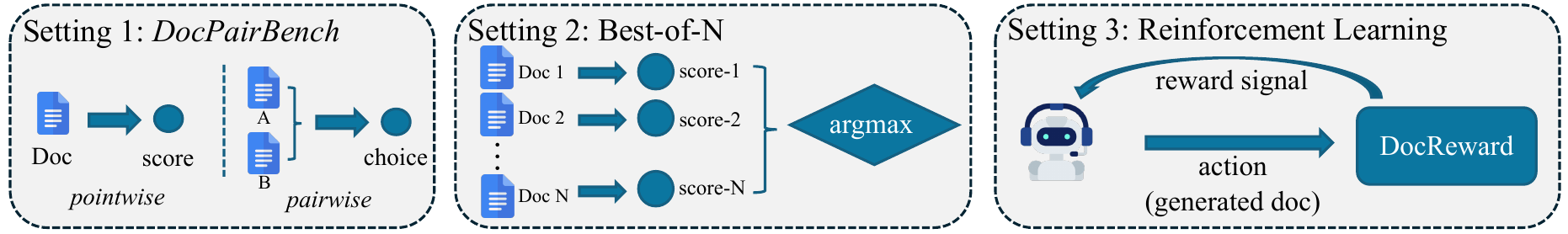}
         \caption{Three evaluation settings for \modelname{}. }
        \label{fig:evaluation_settings}
    \end{figure*}

    We first present the newly proposed benchmark \benchname{} (Section~\ref{sec:docpairbench}) and then conduct intrinsic and extrinsic evaluations of \modelname{} as shown in~\autoref{fig:evaluation_settings} (Section~\ref{sec:results} and~\ref{sec:rl}).
    
\subsection{Human-Annotated \benchname{} Construction} 
\label{sec:docpairbench}
    A subset of the curated documents in Section~\ref{sec:data_construction} is set aside as evaluation documents. To diversify the evaluation samples, we consider the following six types of documents using the method described in Section~\ref{sec:expand_documents_via_agent}. Four of them are obtained via the \textit{Textual Content to Document} agent, which generates DOCX documents using different LLMs ($i.e.$, GPT-4o, OpenAI o1, Claude Sonnet 4, and GPT-5). One type comes from the \textit{Refinement for Better Structure and Style} agent, where GPT-5 is employed to refine synthesized documents. The last type consists of the curated human-authored documents. Together, these six types constitute the origins of samples in our benchmark.
    For each set of documents sharing the same content but differing in structure and style, human experts meticulously rank their quality based on structure and style. To facilitate model evaluation, these ranked relationships are converted into a total of \benchpaircount{} comparison pairs, each consisting of two documents and a binary label indicating the preferred one. To ensure the quality of human annotation, we evaluate consistency among human annotators using Cohen's Kappa and observe a value of 0.834. 
    The details of annotation protocol and reliability validation are presented in Appendix \ref{sec:human_annotation_guidelines_and_reliability}.

\subsection{Results on \benchname{}}
\label{sec:results}
    \begin{table*}[t]
\centering

\caption{Accuracy of Models on the \benchname{} benchmark across ten domains: Government (Gov.), Education (Edu.), Non-Profit Organization (NPO), Medical (Med.), Science (Sci.), Legal (Leg.), Business (Bus.), Academic (Acad.), Technology (Tech.), and Other (Oth.). }
\small
{
\setlength{\tabcolsep}{4pt}

\begin{tabular}{cccccccccccc}
\toprule
\multicolumn{1}{c}{Models} & \multicolumn{1}{c}{Gov.} & \multicolumn{1}{c}{Edu.} & \multicolumn{1}{c}{NPO} & \multicolumn{1}{c}{Med.} & \multicolumn{1}{c}{Sci.} & \multicolumn{1}{c}{Leg.} & \multicolumn{1}{c}{Bus.} & \multicolumn{1}{c}{Acad.} & \multicolumn{1}{c}{Tech.} & \multicolumn{1}{c}{Oth.} & \multicolumn{1}{l}{Overall} \\
\midrule
\multicolumn{12}{c}{\textbf{Pairwise Setting}} \\
\midrule
Qwen2.5-VL-3B\citep{bai2025qwen25vl}        & 55.1                     & 61.5                     & 48.4                    & 50.0                     & 50.0                     & 47.0                     & 56.9                     & 47.3                      & 55.5                      & 60.5                     & 54.4                        \\
Qwen2.5-VL-7B\citep{bai2025qwen25vl}        & 57.7                     & 61.5                     & 53.9                    & 53.8                     & 54.0                     & 59.0                     & 56.0                     & 52.7                      & 49.1                      & 58.1                     & 56.1                        \\
GPT-4o\citep{hurst2024gpt4o}                 & 61.9                     & 57.7                     & 75.8                    & 66.3                     & 53.0                     & 62.0                     & 71.6                     & 61.8                      & 66.4                      & 61.4                     & 63.3                        \\
Claude Sonnet 4\citep{anthropic2025claude4}      & 62.9                     & 69.2                     & 75.8                    & 61.3                     & 46.0                     & 60.0                     & 75.2                     & 56.4                      & 65.5                      & 62.9                     & 63.0                        \\
GPT-5\citep{openai2025gpt5}                 & 70.2                     & 69.2                     & 69.2                    & 68.8                     & 57.0                     & 69.0                     & 76.2                     & 63.6                      & 70.0                      & 69.8                     & 68.9                        \\
\midrule
\multicolumn{12}{c}{\textbf{Pointwise Setting}} \\ \midrule
Qwen2.5-VL-3B\citep{bai2025qwen25vl}        & 38.4                     & 23.1                     & 36.3                    & 30.0                     & 25.0                     & 27.0                     & 32.1                     & 39.1                      & 38.2                      & 30.2                     & 33.5                        \\
Qwen2.5-VL-7B\citep{bai2025qwen25vl}        & 49.4                     & 34.6                     & 46.2                    & 41.3                     & 35.0                     & 52.0                     & 49.5                     & 31.8                      & 48.2                      & 48.5                     & 46.0                        \\
GPT-4o\citep{hurst2024gpt4o}                 & 60.1                     & 46.2                     & 58.2                    & 56.3                     & 34.0                     & 54.0                     & 58.7                     & 50.9                      & 58.2                      & 53.9                     & 54.9                        \\
Claude Sonnet 4\citep{anthropic2025claude4}      & 59.8                     & 50.0                     & 56.0                    & 60.0                     & 48.0                     & 63.0                     & 48.6                     & 50.0                      & 50.9                      & 49.7                     & 54.2                        \\
GPT-5\citep{openai2025gpt5}                 & 69.7                     & 80.8                     & 70.3                    & 71.3                     & 49.0                     & 72.0                     & 75.2                     & 60.0                      & 64.6                      & 66.6                     & 67.7                        \\
\midrule
DocReward-3B (Ours)         & 87.2                     & 76.9                     & 80.2                    & 75.0                     & \textbf{69.0}            & 85.0                     & \textbf{87.2}            & 79.1                      & 74.6                      & \textbf{77.3}            & 80.6                        \\
DocReward-7B (Ours)    & \textbf{89.3}            & \textbf{92.3}            & \textbf{86.8}           & \textbf{83.8}            & 67.0                     & \textbf{87.0}            & 84.4                     & \textbf{80.9}             & \textbf{76.4}             & 76.7                     & \textbf{82.3}              
\\ \bottomrule

\end{tabular}
}

\label{tab:main_exp}
\end{table*}


    As presented in~\autoref{tab:main_exp}, we evaluate methods under two settings, namely \textit{Pairwise} and \textit{Pointwise}. In the pairwise setting, models are given a document pair and asked to select the one with better structure and style, while in the pointwise setting, documents are scored independently. Performance is measured by accuracy with respect to human annotations. On \benchname{}, \modelname{}-3B and \modelname{}-7B achieve substantial improvements over strong baselines including Qwen2.5-VL, GPT-4o, Claude Sonnet 4, and GPT-5. In particular, \modelname{}-7B attains an overall human preference accuracy of 82.3 percent, surpassing the strongest closed-source model (GPT-5, pointwise) by 14.6 points. These results indicate that \modelname{} captures structural and stylistic quality signals that existing language models often overlook.

\subsection{Improving Document Generation with \textbf{\modelname{}}}
\label{sec:rl}
    To demonstrate the utility of \modelname{}, we conduct two complementary experiments.

\begin{table*}[]
\begin{minipage}[]{0.35\linewidth}
\centering
\caption{Best-of-N evaluation results. \modelname{} shows utility for professional document generation. }
\small
{
\setlength{\tabcolsep}{4.5pt}
    \begin{tabular}{cccc}
    \toprule
    Rewards & Win  & Lose & Tie  \\ \midrule
    Random    & 24.6 & 66.2 & 9.2  \\
    GPT-5     & 37.7 & 40.0   & 22.3 \\
    \modelname{} & 60.8 & 16.9 & 22.3 \\ \bottomrule
    \end{tabular}
}
\label{tab:best_of_N}
\end{minipage}
\hfill
\begin{minipage}[]{0.62\linewidth} 
\centering
\caption{Results of the reinforcement learning experiments. ``Score'' denotes the sigmoid-normalized \modelname{} score. ``Rank'' denotes the average ranking among the listed methods.}
\small
{
\setlength{\tabcolsep}{4.5pt}
\begin{tabular}{lcccc}
\toprule
                      Reward Types                      & Success$\uparrow$ & ROUGE-L$\uparrow$ & Score$\uparrow$ & Rank$\downarrow$ \\ \midrule
Qwen2.5-Coder                               & 30.0         & 20.61   & 0.0663             & 4.58          \\
+ rule                           & 98.0         & 97.94   & 0.1785             & 4.06          \\
+ DocReward               & \textbf{100.0}        & \textbf{97.95}   & \textbf{0.3046}             & \textbf{2.84}          \\ \midrule
GPT-4o                                      & 52.0         & 48.73   & 0.2682             & 3.18          \\
+ rule              & 66.0         & 62.15   & 0.3189             & 2.70          \\
+ DocReward & \textbf{78.0}         & \textbf{74.33}   & \textbf{0.4486}             & \textbf{2.02}          \\ \bottomrule
\end{tabular}
}
\label{tab:rl_exp}
\end{minipage}

\end{table*}

    \noindent\textbf{Best-of-N.} 
    A document agent generates $N$ candidates from the same textual content, and a reward model selects the best one based on its score. We compare three reward models including ``Random'', GPT-5 and \modelname{} where human annotators rank the selected outputs by structure and style. As shown in \autoref{tab:best_of_N}, \modelname{} consistently outperforms the baselines, achieving a win rate of 60.8\% against ``Random'' and GPT-5 reward models, indicating that its reward better aligns with human preferences. This evaluation shows that integrating \modelname{} into a document agent improves output quality without modifying the agent itself. Details are provided in Appendix~\ref{appendix:best_of_n_anntoation_detail}.
    
    \noindent\textbf{Reinforcement Learning.}
    We aim to enhance document generation agentic workflows that take plain text as input and generate professional documents. We consider two kinds of rewards: 1) $\mathcal{R}_{\texttt{rule}}$ that penalizes documents that either result from invalid \texttt{python-docx} code or differ from the input text after execution. Specifically, if the code executes successfully, then $\mathcal{R}_{\texttt{rule}} = ROUGE(doc\_{ori}, doc\_{gen})$; else $\mathcal{R}_{\texttt{rule}}$ will be zero.  2) $\mathcal{R}_{\texttt{DocReward}}$ that penalizes documents with poor structure and style. Overall, the total reward is defined as:
    
    \begin{equation}
    \mathcal{R} = \mathcal{R}_{\texttt{rule}} + \alpha \cdot \mathbb{I}_{\text{rule}} \cdot \sigma(\mathcal{R}_{\texttt{DocReward}}),\label{eq:reward}
    \end{equation}
    
    where $\alpha$ is a hyperparameter to balance the rewards, $\sigma(\cdot)$ is the $\mathrm{Sigmoid}$ operation to regularize the value range of $\modelname{}$ to $(0, 1)$, and $\mathbb{I}_{\text{rule}}$ represents whether $\mathcal{R}_{\texttt{rule}}$ is larger than a threshold.
    
\begin{figure*}[t]
        \centering
        \includegraphics[width=0.95\linewidth]{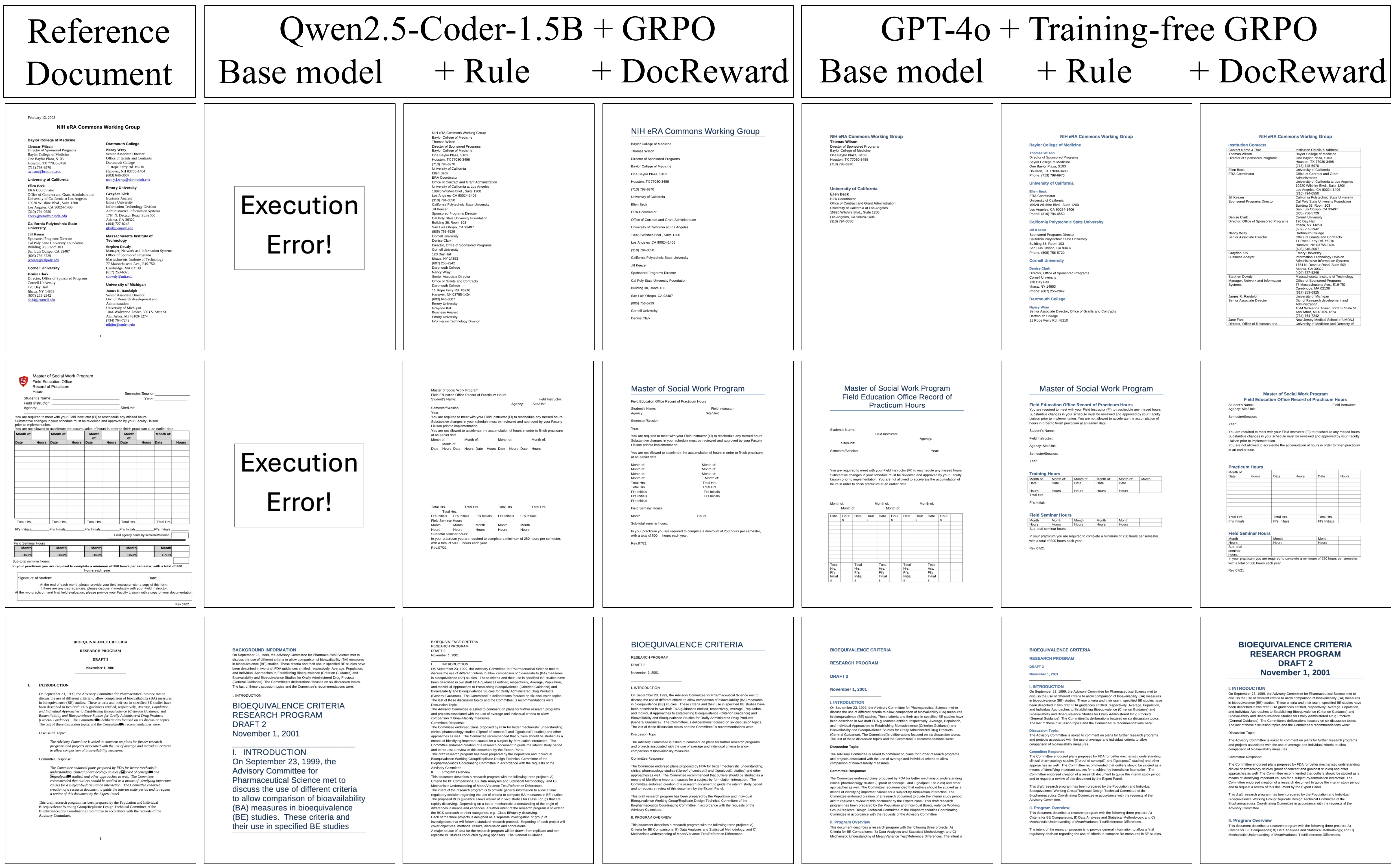}
         \caption{Visualization of documents generated by various methods. “Execution error” denotes instances where the generated code encountered runtime failures, preventing the final document from being rendered.}
        \label{fig:visualization_case_rl}
    \end{figure*}

    For open-source models, we adopt GRPO~\citep{shao2024deepseekmath} as the RL algorithm, while employing training-free GRPO~\citep{trainingfreegrpo} for closed-source models.\footnote{Training-free GRPO is a lightweight approach without model parameter updates.} To evaluate the quality of RL-trained models, human annotators rank documents produced by six model variants based on the professionalism of structure and style. As shown in~\autoref{tab:rl_exp}, rule-based rewards substantially improve document generation success rate for both Qwen2.5-Coder and GPT-4o. Further incorporating DocReward consistently enhances performance across both open- and closed-source models, yielding higher success rates, improved ROUGE-L scores, and better human rankings. These results demonstrate that \modelname{} serves as an effective and generalizable reward model for professional structure and style.~\autoref{fig:visualization_case_rl} presents a visualization of documents generated by different methods.

\subsection{Robustness Test}

\begin{table*}[]
\begin{minipage}[b]{0.38\linewidth}
\centering
\caption{Out-of-domain (OOD) experiment. \modelname{} generalizes effectively to unseen domains.}
\small
{
    \setlength{\tabcolsep}{4.5pt}
    \begin{tabular}{ccc}
        \toprule
        \textbf{Model} & \textbf{ID} & \textbf{OOD} \\
        \midrule
        Qwen2.5-VL-3B   & 33.5                          & 30.0                              \\
        Qwen2.5-VL-7B   & 46.0                          & 48.1                              \\
        GPT-4o          & 54.9                          & 53.8                              \\
        Claude Sonnet 4 & 54.2                          & 49.1                              \\
        GPT-5           & 67.7                          & 68.4                              \\
        DocReward-3B    & 80.6                          & 76.3                              \\
        DocReward-7B    & \textbf{82.3}                          & \textbf{77.5}       \\
        \bottomrule
    \end{tabular}
}

    \label{tab:ood}
\end{minipage}
\hfill
\begin{minipage}[b]{0.6\linewidth} 
\centering
\caption{Cross-lingual robustness evaluation. \textbf{$\Delta$} denotes the performance gap between English and non-English scenarios. \modelname{} demonstrates strong cross-lingual robustness. }
\small
{
\setlength{\tabcolsep}{4.5pt}
\begin{tabular}{ccccccc}
    \toprule
\textbf{Model} & \textbf{FR} & \textbf{ES} & \textbf{RU} & \textbf{Non-EN} & \textbf{EN} & \textbf{$\Delta$}
 \\
        \midrule
Qwen2.5-VL-3B   & 35.0                       & 33.8                        & 22.5                        & 30.4                             & 33.5                        & -3.1 \\
Qwen2.5-VL-7B   & 48.8                       & 42.5                        & 27.5                        & 39.6                             & 46.0                        & -6.4 \\
GPT-4o          & 47.5                       & 52.5                        & 42.5                        & 47.5                             & 54.9                        & -7.4 \\
Claude Sonnet 4 & 57.5                       & 51.3                        & 42.5                        & 50.4                             & 54.2                        & -3.8 \\
GPT-5           & 57.5                       & 76.3                        & 47.5                        & 60.4                             & 67.7                        & -7.3 \\
DocReward-3B    & 77.5                       & 82.5                        & 72.5                        & 77.5                             & 80.6                        & -3.1 \\
DocReward-7B    & 78.8                       & 88.8                        & 66.3                        & 77.9                             & 82.3                        & -4.4                       \\    
        \bottomrule
    \end{tabular}
}
\label{tab:crosslingual}
\end{minipage}

\end{table*}
    
    \noindent\textbf{Out-of-Domain Evaluation.} To evaluate cross-domain robustness, certain domains are reserved exclusively as test sets rather than being used for training. \autoref{tab:ood} reports the out-of-domain results of different models. Firstly, \modelname{}-7B (77.5) remains superior to all baseline models, including the closed-source model GPT-5 (68.4). This trend is consistent with the in-domain results. Secondly, the performance of \modelname{}-7B decreases by merely 4.8 percentage points when transitioning from in-domain to out-of-domain evaluations. Such a small performance gap indicates that \modelname{} generalizes effectively to unseen domains.

    \noindent\textbf{Cross-Lingual Robustness.} To evaluate the cross-lingual robustness of \modelname{} trained on English data, we conduct evaluations in French, Spanish, and Russian, with results shown in~\autoref{tab:crosslingual}. Firstly, in non-English scenarios, \modelname{}-7B model achieves a score of 77.9, substantially outperforming all baseline models. Secondly, all models, including the baselines, exhibit performance degradation in non-English settings. Notably, the performance drop of \modelname{} ($-4.4$) is even smaller than that of closed-source models GPT-4o ($-7.4$) and GPT-5 ($-7.3$), indicating that \modelname{} demonstrates strong cross-lingual robustness.

\subsection{Visualization of Attention Maps}
\begin{figure*}[]
    \centering
    \begin{minipage}[b]{0.26\linewidth}
        \includegraphics[width=1\textwidth]{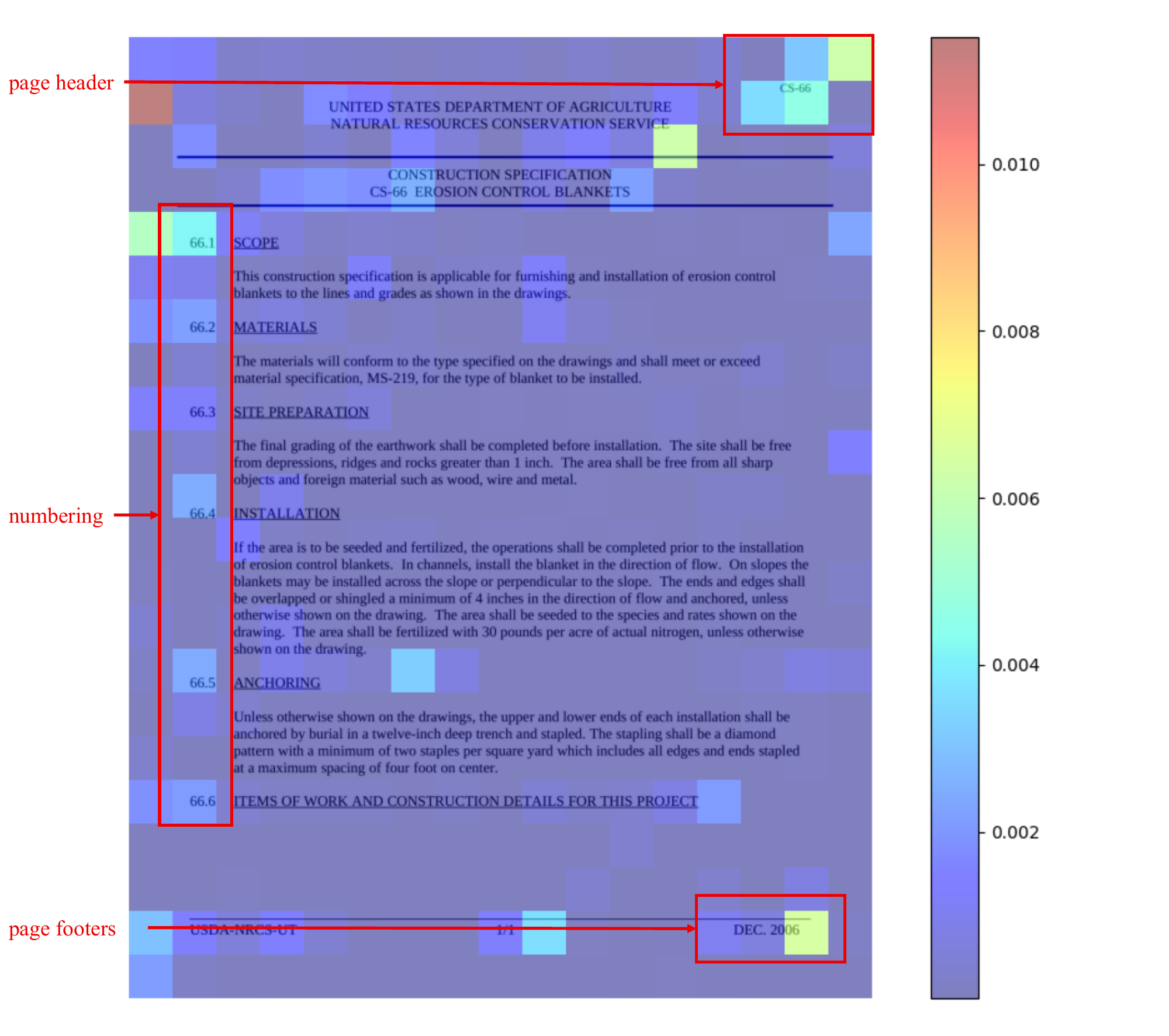}
        \subcaption{}
        \label{fig:attn_map_a}
    \end{minipage}
    \quad
    \begin{minipage}[b]{0.26\linewidth} 
        \includegraphics[width=1\textwidth]{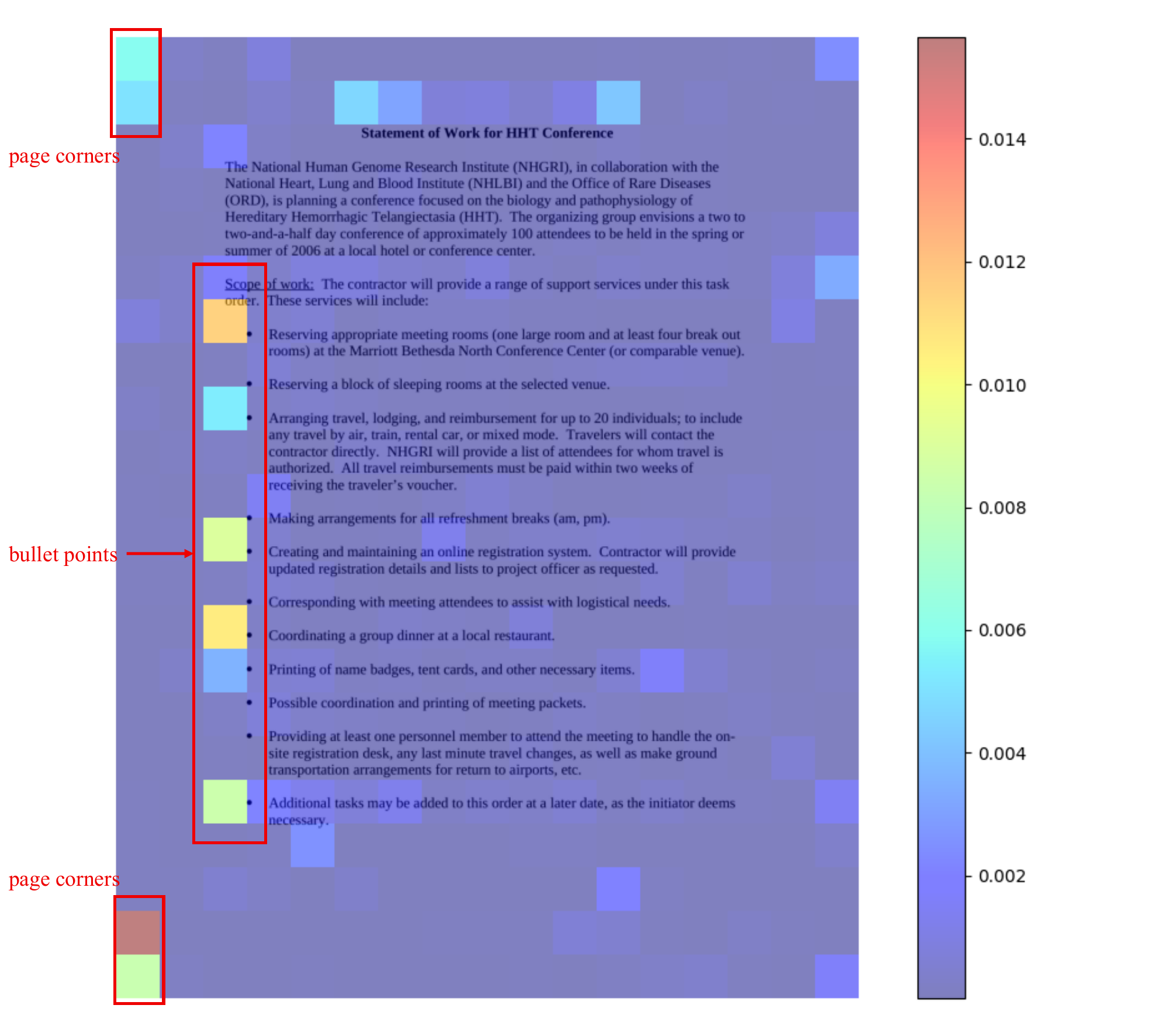}
        \subcaption{}
        \label{fig:attn_map_b}
    \end{minipage}
    \quad
    \begin{minipage}[b]{0.26\linewidth} 
        \includegraphics[width=1\textwidth]{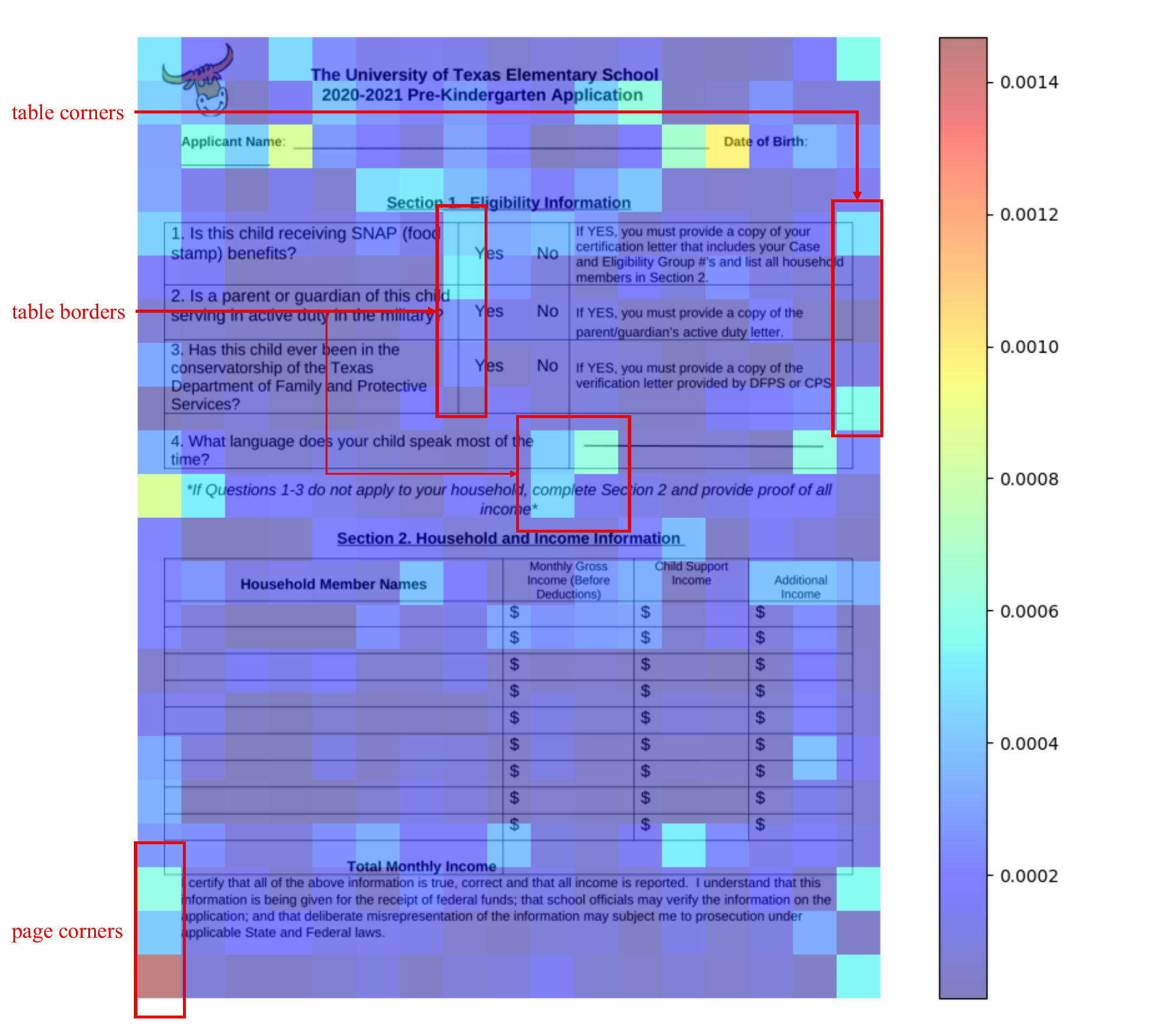}
        \subcaption{}
        \label{fig:attn_map_c}
    \end{minipage}
    \caption{Visualization of attention maps. \modelname{} captures structural and stylistic elements, such as headings, alignment, and whitespace, in its evaluation of document professionalism.}
    \label{fig:attention_map}
    \end{figure*}

    To understand \modelname{}’s decision-making process, we analyze attention maps. As shown in~\autoref{fig:attention_map}, the model attends to structural and formatting cues when assessing document professionalism. Specifically, attention focuses on headings and numbering (\autoref{fig:attn_map_a}), page headers and footers (e.g., “CS-66”, “DEC. 2006”), bullet points (\autoref{fig:attn_map_b}), and table grids (\autoref{fig:attn_map_c}), reflecting sensitivity to structural clarity, formatting consistency, alignment, and readability. Additionally, attention to page corners suggests that uniform margins and balanced whitespace are important indicators of professional layout design.
        
\subsection{Case Study}
    
\begin{figure*}[]
    \centering
    \begin{minipage}[b]{0.26\linewidth} 
        \includegraphics[width=1\textwidth]{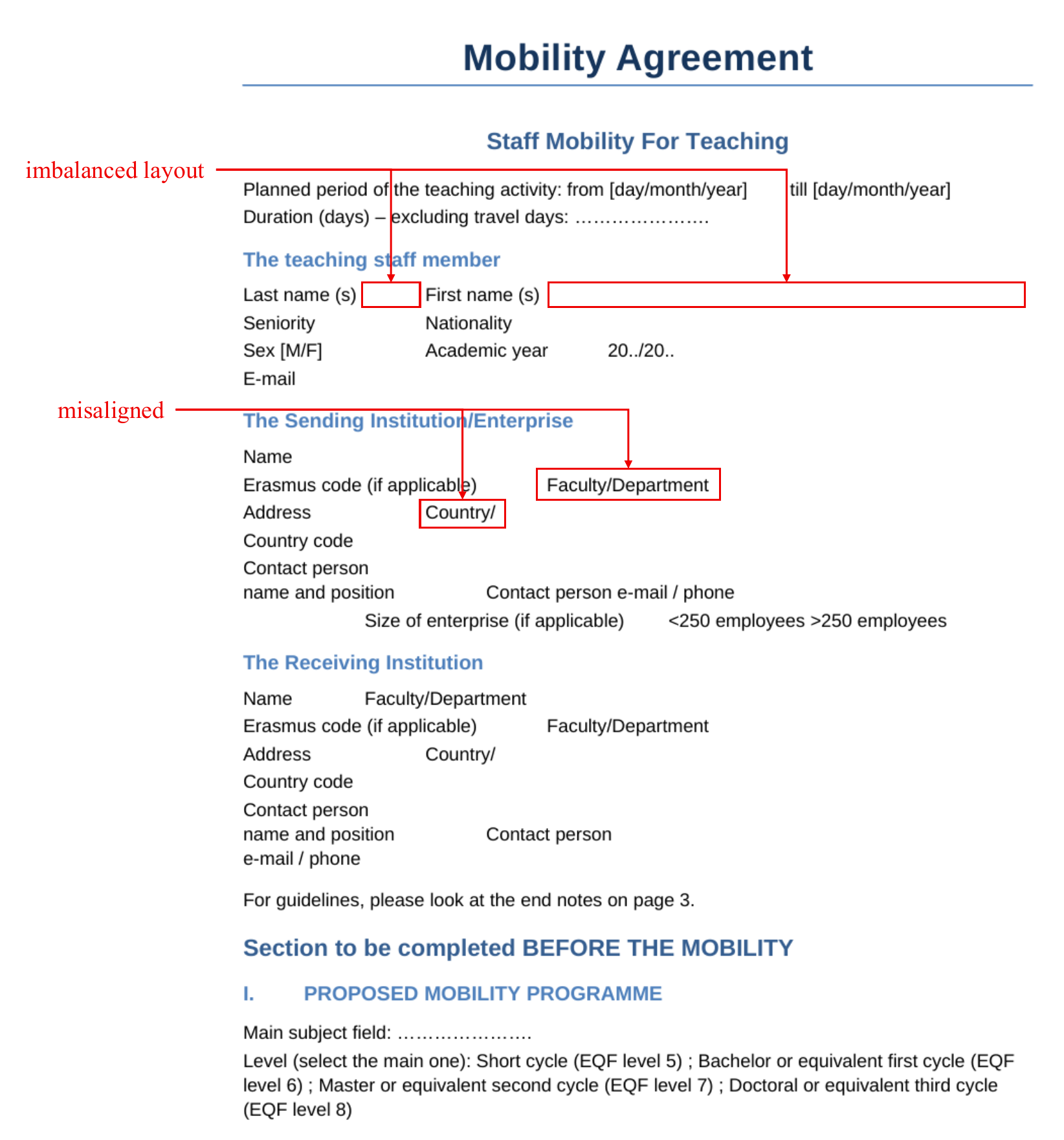}
        \subcaption{score: 1.21}
        \label{fig:case_study_a}
    \end{minipage}
    \quad
    \begin{minipage}[b]{0.26\linewidth} 
        \includegraphics[width=1\textwidth]{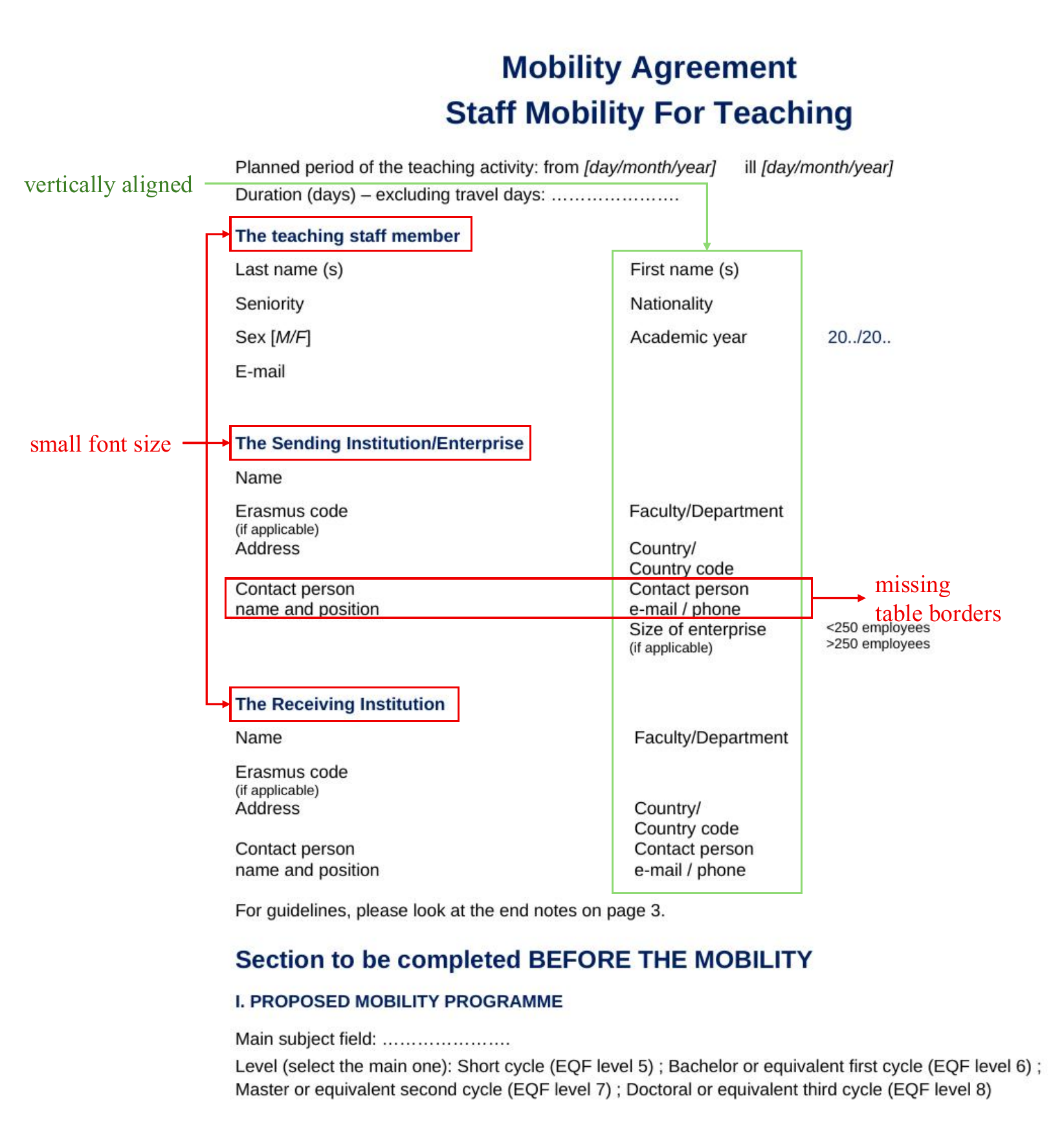}
        \subcaption{score: 2.11}
        \label{fig:case_study_b}
    \end{minipage}
    \quad
    \begin{minipage}[b]{0.26\linewidth} 
        \includegraphics[width=1\textwidth]{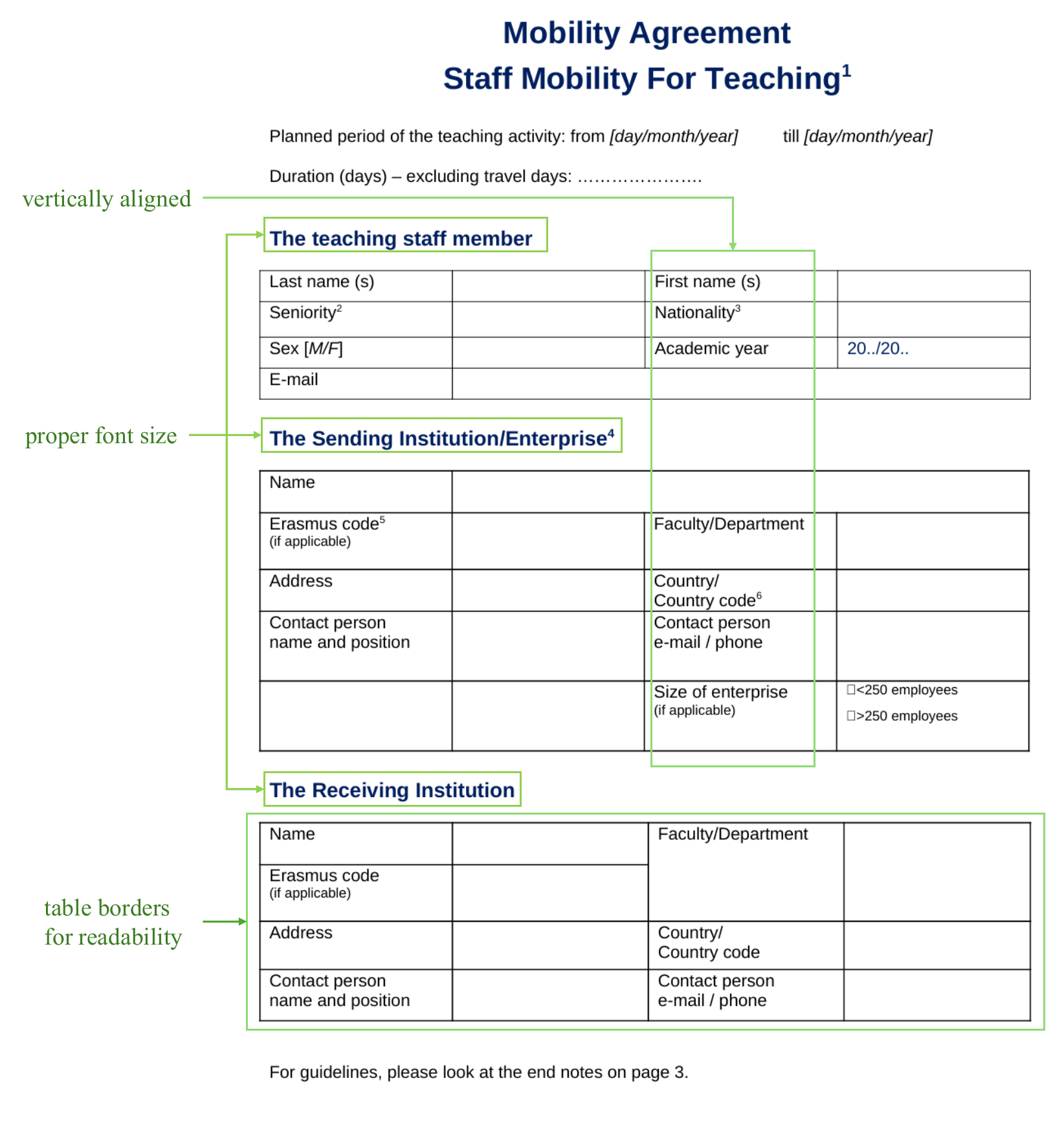}
        \subcaption{score: 5.34}
        \label{fig:case_study_c}
    \end{minipage}
    \caption{Case study: \modelname{}'s scores reflect structural and stylistic professionalism.}
    \label{fig:case_study}
    \end{figure*}

    In~\autoref{fig:case_study}, we present a case study on documents with identical textual content but different structures and styles. In case (a), the allocation of whitespace is ineffective, with insufficient space allocated for \textit{Last Name} and excessive space for \textit{First Name}, leading to an imbalanced layout. Key fields such as \textit{Faculty/Department}, \textit{Country}, and \textit{Country Code} are not vertically aligned, causing a cluttered and disorganized layout. This poor alignment and inconsistent spacing result in a low score of 1.21 from \modelname{}. Case (b) adopts a table-like arrangement, but the level-1 heading "\textit{The teaching staff member}" is too small and does not stand out from the body text, diminishing its impact. Additionally, the lack of borders around input fields makes it hard to locate items, resulting in a moderate score of 2.11. Case (c) provides a clear and well-structured layout, with appropriately sized headings and improved readability, earning the highest rating of 5.34. These results show that \modelname{} effectively captures document professionalism in structure and style. 

\section{Related Work}

\paragraph{Aesthetic and Professionalism Assessment.}
In graphic design, {AesthetiQ}~\citep{patnaik2025aesthetiq} trains an MLLM for layout prediction that uses MLLM’s aesthetic preferences for DPO over graphic layouts, while diffusion-based methods such as {LACE}~\citep{chen2024towards} introduce differentiable constraints to directly optimize layout attributes. 
For web interfaces, {Calista}~\citep{DELITZAS2023103019} uses explicit ratings and pairwise comparisons to model visual appeal, showing high correlation to human perception. 
Additionally, photo aesthetics are modeled using layout-aware CNNs such as {A-Lamp}~\citep{DBLP:conf/cvpr/MaLC17}, and similar techniques extend to video~\citep{liu2023ai}.  
These studies focus on images or UI interfaces rather than multi-page documents, where professionalism depends on both structure and style.

\paragraph{Document AI.}
Document AI research mainly targets semantic parsing and content understanding. 
Models such as LayoutLM~\citep{xu2020layoutlm} and ReLayout~\citep{jiang2025relayout}, along with OCR-based pipelines~\citep{subramani2020survey}, identify logical elements such as headings, tables, and semantic groups to support information extraction and classification. 
Recent work also explores automatic document or layout generation~\citep{lin2023layoutprompter,tang2023layoutnuwa,tian2025relayout}, but evaluation has primarily been limited to content correctness or basic formatting. 
As a result, the assessment of document professionalism—particularly visual structure and style—remains largely unexplored.

\paragraph{Preference Learning and Reward Models.}
A major challenge in professionalism assessment is acquiring feedback signals that reflect human judgment. 
Preference-based reward modeling addresses this issue by training on pairwise comparisons to approximate preferences, forming the basis of alignment methods like RLHF~\citep{stiennon2020learning} and DPO~\citep{dpo}. 

\section{Conclusion}
In this paper, we introduced \modelname{}, a document reward model designed to assess structural and stylistic professionalism. To enable professionalism assessment without being influenced by textual quality, we propose a \textit{textual-quality-agnostic} framework for document reward modeling. Then, a multi-domain preference dataset, \dataname{}, consisting of 117K paired documents, is constructed to enable the training of \modelname{} using the Bradley-Terry loss. Comprehensive experiments demonstrate that \modelname{} not only outperforms GPT-5 in assessing structure and style in the same setting, but also serves as an effective reward model in reinforcement learning for agentic workflows.


\bibliographystyle{unsrt}
\bibliography{sections/custom}

\begin{thebibliography}{10}

\bibitem{peng2023copilot}
Sida Peng, Eirini Kalliamvakou, Peter Cihon, and Mert Demirer.
\newblock The impact of ai on developer productivity: Evidence from github copilot.
\newblock {\em arXiv preprint arXiv:2302.06590}, 2023.

\bibitem{anthropic2025claudecode}
Anthropic.
\newblock Claude code: Best practices for agentic coding.
\newblock \url{https://www.anthropic.com/engineering/claude-code-best-practices}, April 2025.
\newblock Accessed: 2025-09-24.

\bibitem{hong2024metagpt}
Sirui Hong, Mingchen Zhuge, Jonathan Chen, Xiawu Zheng, Yuheng Cheng, Jinlin Wang, Ceyao Zhang, Zili Wang, Steven Ka~Shing Yau, Zijuan Lin, Liyang Zhou, Chenyu Ran, Lingfeng Xiao, Chenglin Wu, and J{\"u}rgen Schmidhuber.
\newblock Meta{GPT}: Meta programming for a multi-agent collaborative framework.
\newblock In {\em The Twelfth International Conference on Learning Representations}, 2024.

\bibitem{Xiao_2025_CVPR}
Shitao Xiao, Yueze Wang, Junjie Zhou, Huaying Yuan, Xingrun Xing, Ruiran Yan, Chaofan Li, Shuting Wang, Tiejun Huang, and Zheng Liu.
\newblock Omnigen: Unified image generation.
\newblock In {\em Proceedings of the IEEE/CVF Conference on Computer Vision and Pattern Recognition (CVPR)}, pages 13294--13304, June 2025.

\bibitem{zheng2025deepeyes}
Ziwei Zheng, Michael Yang, Jack Hong, Chenxiao Zhao, Guohai Xu, Le~Yang, Chao Shen, and Xing Yu.
\newblock Deepeyes: Incentivizing" thinking with images" via reinforcement learning.
\newblock {\em arXiv preprint arXiv:2505.14362}, 2025.

\bibitem{Marsili2025Visual_agentic_ai}
Damiano Marsili, Rohun Agrawal, Yisong Yue, and Georgia Gkioxari.
\newblock Visual agentic ai for spatial reasoning with a dynamic api.
\newblock In {\em Proceedings of the Computer Vision and Pattern Recognition Conference}, pages 19446--19455, 2025.

\bibitem{yan2025mathagent}
Yibo Yan, Shen Wang, Jiahao Huo, Philip~S. Yu, Xuming Hu, and Qingsong Wen.
\newblock {M}ath{A}gent: Leveraging a mixture-of-math-agent framework for real-world multimodal mathematical error detection.
\newblock In Georg Rehm and Yunyao Li, editors, {\em Proceedings of the 63rd Annual Meeting of the Association for Computational Linguistics (Volume 6: Industry Track)}, pages 69--82, Vienna, Austria, July 2025. Association for Computational Linguistics.

\bibitem{xie2024travelplanner}
Jian Xie, Kai Zhang, Jiangjie Chen, Tinghui Zhu, Renze Lou, Yuandong Tian, Yanghua Xiao, and Yu~Su.
\newblock Travelplanner: A benchmark for real-world planning with language agents.
\newblock {\em arXiv preprint arXiv:2402.01622}, 2024.

\bibitem{openai2025deepresearch}
OpenAI.
\newblock Introducing deep research.
\newblock \url{https://openai.com/index/introducing-deep-research/}, February 2025.
\newblock Accessed: 2025-09-24.

\bibitem{openmanus2025}
Xinbin Liang, Jinyu Xiang, Zhaoyang Yu, Jiayi Zhang, Sirui Hong, Sheng Fan, Xiao Tang, Bang Liu, Yuyu Luo, and Chenglin Wu.
\newblock Openmanus: An open-source framework for building general ai agents, 2025.

\bibitem{qwen2025deepresearch}
Alibaba Cloud~/ Qwen.
\newblock Deep research — qwen.
\newblock \url{https://chat.qwen.ai/?inputFeature=deep_research}, 2025.
\newblock Accessed: 2025-09-24.

\bibitem{dvivedi2024code_documentation}
Shubhang~Shekhar Dvivedi, Vyshnav Vijay, Sai Leela~Rahul Pujari, Shoumik Lodh, and Dhruv Kumar.
\newblock A comparative analysis of large language models for code documentation generation.
\newblock In {\em Proceedings of the 1st ACM international conference on AI-powered software}, pages 65--73, 2024.

\bibitem{bradley1952rank}
Ralph~Allan Bradley and Milton~E Terry.
\newblock Rank analysis of incomplete block designs: I. the method of paired comparisons.
\newblock {\em Biometrika}, 39(3/4):324--345, 1952.

\bibitem{ouyang2022training}
Long Ouyang, Jeffrey Wu, Xu~Jiang, Diogo Almeida, Carroll Wainwright, Pamela Mishkin, Chong Zhang, Sandhini Agarwal, Katarina Slama, Alex Ray, et~al.
\newblock Training language models to follow instructions with human feedback.
\newblock {\em Advances in neural information processing systems}, 35:27730--27744, 2022.

\bibitem{hurst2024gpt4o}
Aaron Hurst, Adam Lerer, Adam~P Goucher, Adam Perelman, Aditya Ramesh, Aidan Clark, AJ~Ostrow, Akila Welihinda, Alan Hayes, Alec Radford, et~al.
\newblock Gpt-4o system card.
\newblock {\em arXiv preprint arXiv:2410.21276}, 2024.

\bibitem{openai2025gpt5}
OpenAI.
\newblock Gpt-5 system card.
\newblock \url{https://openai.com/index/gpt-5-system-card/}, August 2025.
\newblock Accessed: 2025-09-24.

\bibitem{hui2024qwen25coder}
Binyuan Hui, Jian Yang, Zeyu Cui, Jiaxi Yang, Dayiheng Liu, Lei Zhang, Tianyu Liu, Jiajun Zhang, Bowen Yu, Keming Lu, et~al.
\newblock Qwen2.5-coder technical report.
\newblock {\em arXiv preprint arXiv:2409.12186}, 2024.

\bibitem{garfinkel2009govdocs}
Simson Garfinkel, Paul Farrell, Vassil Roussev, and George Dinolt.
\newblock Bringing science to digital forensics with standardized forensic corpora.
\newblock {\em digital investigation}, 6:S2--S11, 2009.

\bibitem{davies2022napierone}
Simon~R Davies, Richard Macfarlane, and William~J Buchanan.
\newblock Napierone: A modern mixed file data set alternative to govdocs1.
\newblock {\em Forensic Science International: Digital Investigation}, 40:301330, 2022.

\bibitem{openai2024o1system}
OpenAI.
\newblock Openai o1 system card.
\newblock \url{https://openai.com/index/openai-o1-system-card/}, December 2024.
\newblock Updated: December 5, 2024. Accessed: 2025-09-24.

\bibitem{anthropic2025claude4}
Anthropic.
\newblock Introducing claude 4.
\newblock \url{https://www.anthropic.com/news/claude-4}, May 2025.
\newblock Accessed: 2025-09-24.

\bibitem{bai2025qwen25vl}
Shuai Bai, Keqin Chen, Xuejing Liu, Jialin Wang, Wenbin Ge, Sibo Song, Kai Dang, Peng Wang, Shijie Wang, Jun Tang, Humen Zhong, Yuanzhi Zhu, Mingkun Yang, Zhaohai Li, Jianqiang Wan, Pengfei Wang, Wei Ding, Zheren Fu, Yiheng Xu, Jiabo Ye, Xi~Zhang, Tianbao Xie, Zesen Cheng, Hang Zhang, Zhibo Yang, Haiyang Xu, and Junyang Lin.
\newblock Qwen2.5-vl technical report.
\newblock {\em ArXiv}, abs/2502.13923, 2025.

\bibitem{shao2024deepseekmath}
Zhihong Shao, Peiyi Wang, Qihao Zhu, Runxin Xu, Junxiao Song, Xiao Bi, Haowei Zhang, Mingchuan Zhang, YK~Li, Yang Wu, et~al.
\newblock Deepseekmath: Pushing the limits of mathematical reasoning in open language models.
\newblock {\em arXiv preprint arXiv:2402.03300}, 2024.

\bibitem{trainingfreegrpo}
Yuzheng Cai, Siqi Cai, Yuchen Shi, Zihan Xu, Lichao Chen, Yulei Qin, Xiaoyu Tan, Gang Li, Zongyi Li, Haojia Lin, et~al.
\newblock Training-free group relative policy optimization.
\newblock {\em arXiv preprint arXiv:2510.08191}, 2025.

\bibitem{patnaik2025aesthetiq}
Sohan Patnaik, Rishabh Jain, Balaji Krishnamurthy, and Mausoom Sarkar.
\newblock Aesthetiq: Enhancing graphic layout design via aesthetic-aware preference alignment of multi-modal large language models.
\newblock In {\em Proceedings of the Computer Vision and Pattern Recognition Conference}, pages 23701--23711, 2025.

\bibitem{chen2024towards}
Jian Chen, Ruiyi Zhang, Yufan Zhou, and Changyou Chen.
\newblock Towards aligned layout generation via diffusion model with aesthetic constraints.
\newblock In {\em The Twelfth International Conference on Learning Representations, {ICLR} 2024, Vienna, Austria, May 7-11, 2024}. OpenReview.net, 2024.

\bibitem{DELITZAS2023103019}
Alexandros Delitzas, Kyriakos~C. Chatzidimitriou, and Andreas~L. Symeonidis.
\newblock Calista: A deep learning-based system for understanding and evaluating website aesthetics.
\newblock {\em International Journal of Human-Computer Studies}, 175:103019, 2023.

\bibitem{DBLP:conf/cvpr/MaLC17}
Shuang Ma, Jing Liu, and Chang~Wen Chen.
\newblock A-lamp: Adaptive layout-aware multi-patch deep convolutional neural network for photo aesthetic assessment.
\newblock In {\em 2017 {IEEE} Conference on Computer Vision and Pattern Recognition, {CVPR} 2017, Honolulu, HI, USA, July 21-26, 2017}, pages 722--731. {IEEE} Computer Society, 2017.

\bibitem{liu2023ai}
Chang Liu and Han Yu.
\newblock Ai-empowered persuasive video generation: A survey.
\newblock {\em ACM Computing Surveys}, 55(13s):1--31, 2023.

\bibitem{xu2020layoutlm}
Yiheng Xu, Minghao Li, Lei Cui, Shaohan Huang, Furu Wei, and Ming Zhou.
\newblock Layoutlm: Pre-training of text and layout for document image understanding.
\newblock In {\em Proceedings of the 26th ACM SIGKDD international conference on knowledge discovery \& data mining}, pages 1192--1200, 2020.

\bibitem{jiang2025relayout}
Zhouqiang Jiang, Bowen Wang, Junhao Chen, and Yuta Nakashima.
\newblock Relayout: Towards real-world document understanding via layout-enhanced pre-training.
\newblock In Owen Rambow, Leo Wanner, Marianna Apidianaki, Hend Al{-}Khalifa, Barbara~Di Eugenio, and Steven Schockaert, editors, {\em Proceedings of the 31st International Conference on Computational Linguistics, {COLING} 2025, Abu Dhabi, UAE, January 19-24, 2025}, pages 3778--3793. Association for Computational Linguistics, 2025.

\bibitem{subramani2020survey}
Nishant Subramani, Alexandre Matton, Malcolm Greaves, and Adrian Lam.
\newblock A survey of deep learning approaches for ocr and document understanding.
\newblock {\em arXiv preprint arXiv:2011.13534}, 2020.

\bibitem{lin2023layoutprompter}
Jiawei Lin, Jiaqi Guo, Shizhao Sun, Zijiang Yang, Jian-Guang Lou, and Dongmei Zhang.
\newblock Layoutprompter: Awaken the design ability of large language models.
\newblock {\em Advances in Neural Information Processing Systems}, 36:43852--43879, 2023.

\bibitem{tang2023layoutnuwa}
Zecheng Tang, Chenfei Wu, Juntao Li, and Nan Duan.
\newblock Layoutnuwa: Revealing the hidden layout expertise of large language models.
\newblock {\em arXiv preprint arXiv:2309.09506}, 2023.

\bibitem{tian2025relayout}
Jiaxu Tian, Xuehui Yu, Yaoxing Wang, Pan Wang, Guangqian Guo, and Shan Gao.
\newblock Relayout: Integrating relation reasoning for content-aware layout generation with multi-modal large language models.
\newblock {\em arXiv preprint arXiv:2507.05568}, 2025.

\bibitem{stiennon2020learning}
Nisan Stiennon, Long Ouyang, Jeffrey Wu, Daniel Ziegler, Ryan Lowe, Chelsea Voss, Alec Radford, Dario Amodei, and Paul~F Christiano.
\newblock Learning to summarize with human feedback.
\newblock {\em Advances in neural information processing systems}, 33:3008--3021, 2020.

\bibitem{dpo}
Rafael Rafailov, Archit Sharma, Eric Mitchell, Christopher~D Manning, Stefano Ermon, and Chelsea Finn.
\newblock Direct preference optimization: Your language model is secretly a reward model.
\newblock {\em Advances in neural information processing systems}, 36:53728--53741, 2023.

\bibitem{zheng2024llamafactory}
Yaowei Zheng, Richong Zhang, Junhao Zhang, Yanhan Ye, Zheyan Luo, Zhangchi Feng, and Yongqiang Ma.
\newblock Llamafactory: Unified efficient fine-tuning of 100+ language models.
\newblock {\em arXiv preprint arXiv:2403.13372}, 2024.

\end{thebibliography}

\newpage
\appendix


\section{Appendix}

\subsection{Limitations and Broader Impacts}
\label{sec:limitation}
\paragraph{Limitations and Future Work: } A limitation is that \textsc{DocReward} is designed as a scalar reward model, providing only a numerical score to represent the overall structural and stylistic quality of a document. It currently cannot generate natural language feedback explaining why a document is perceived as unprofessional. Therefore, extending \textsc{DocReward} from a scalar model to an interpretable reward model that can produce both quantitative scores and qualitative diagnostic rationales remains a key focus of our future work.

\label{appendix:broader_impacts}
\paragraph{Broader Impacts: } The document reward model proposed in this work advances agentic workflows for documents. It poses no obvious negative societal impact, as the focus is the \textit{textual-quality-agnostic} training framework and methods.

\subsection{Model Implementation Details} 
    \label{appendix:model_impl_detail}
    Our document reward model is built upon the Qwen2.5-VL multimodal architecture, and a regression head is added to predict scalar scores. The maximum input pixels are set to 300,000. It is configured with a maximum context length of 16,000 tokens. Training utilizes the AdamW optimizer with a learning rate of 1e-6 and a batch size of 256 over 3 epochs. All training was conducted on 8 NVIDIA A100 GPUs (80GB). The training code is based on LLaMA-Factory~\citep{zheng2024llamafactory}. $\alpha$ and the threshold of $\mathbb{I}_{\text{rule}}$ are set to 1 and 0.8, respectively.

    
\subsection{Source Documents Expansion}
    \label{appendix:data_construction_detail}
    To ensure that the reward model learns to assess differences in structure and style rather than content, we applied a rigorous filtering process. Using \texttt{python-docx}, we extracted text from pairs of Microsoft Word DOCX documents and computed their word counts. Only synthetic documents with a word count difference of no more than 20 words from the original document and a ROUGE-L score exceeding a threshold (i.e., 0.95) are retained, ensuring comparable content while isolating variation in structure and style. For the constructed training dataset \dataname{}, both GPT-4o and GPT-5 serve as the base models of agents. 

\subsection{Annotation Protocol and Reliability}
\label{sec:human_annotation_guidelines_and_reliability}

\noindent\textbf{Annotation Guidelines.} 
The detailed guidelines for human annotation are presented in~\autoref{box:guideline}.
The annotation guidelines consist of general principles that are formulated in an explicit, objective manner. For instance, extremely narrow margins that produce an almost fully saturated page layout are commonly regarded as unprofessional across different cultural and regional contexts. 

\begin{figure}[htbp] 
\centering
\begin{myverbatimbox}[Guidelines for Human Annotation]
Target:
Each document group contains N documents. Their textual content is the same, but their structure and style differ. The first document in each group is the original human-authored document, which serves as a reference during annotation. Based on the level of professionalism in structure and style across the N documents, the annotator should rank the documents. Note that there may exist cases where human-authored documents are not the best ones.

Annotation Format Example:
For example, for the document group with ID 10655307, suppose the human-annotated professionalism ranking is 1 > 5 > 3 > 2 > 4, where 1 is judged to be the most professionally structured and formatted document, and 4 is the least professional. Then the annotation format should be: “10655307 \t 15324”

Evaluation Criteria:

1. **Layout and Design**: 
   - Consistent formatting and spacing
   - Proper use of headings, subheadings, and other structures, and proper hierarchy (e.g., long paragraphs should use body text style rather than heading styles, and headings should not be formatted as body text)
   - Appropriate margins and white space usage (e.g., page margins or table column widths that are excessively wide or narrow are not appropriate)
2. **Readability and Typography**:
   - Consistent and appropriate font choices
   - Proper Text size (e.g., overly large or overly small text is not suitable)   
   - Appropriate line spacing and clear paragraph structure
   - Proper Text alignment 
3. **Professional Standards**:
   - Document structure and organization
   - Use of professional elements (headers, footers, page numbers)
   - Consistency across pages (if multiple pages provided)
4. **Visual Elements**:
   - Quality and placement of images, tables, or charts
   - Integration of visual elements with text
   - Professional presentation of data

\end{myverbatimbox}    
\caption{Detailed guideline for human annotation.}
\label{box:guideline}
\end{figure}

\noindent\textbf{Independence from annotators' cultural and professional backgrounds.} 
The annotation was performed by three Ph.D. students with complementary expertise in computer science/math, marketing, and design.
We measured inter-annotator reliability using Cohen's Kappa; the results are shown in~\autoref{tab:kappa}. 
The high agreement indicates that the annotations follow clear, well-defined rules that do not depend on the annotators' professional training or cultural background, demonstrating the guidelines' generality and objectivity.

\begin{table}[b]
  \centering
  \caption{Cohen's Kappa among annotators.}
  \begin{tabular}{ccccc}
    \toprule
    Annotator ID & {1} & {2} & {3} & {Average} \\
    \midrule
    1 & {-}   & 0.8340 & 0.8092 & 0.8215 \\
    2 & 0.8340 & {-}   & 0.8590 & 0.8465 \\
    3 & 0.8092 & 0.8590 & {-}   & 0.8341 \\
    \midrule
    Average     & 0.8215 & 0.8465 & 0.8341 & 0.8340 \\
    \bottomrule
  \end{tabular}
  \label{tab:kappa}
\end{table}

\subsection{Document Types Distribution}
    \label{appendix:data_statistics}
    
\begin{figure}[]
        \centering
        \includegraphics[width=0.6\textwidth]{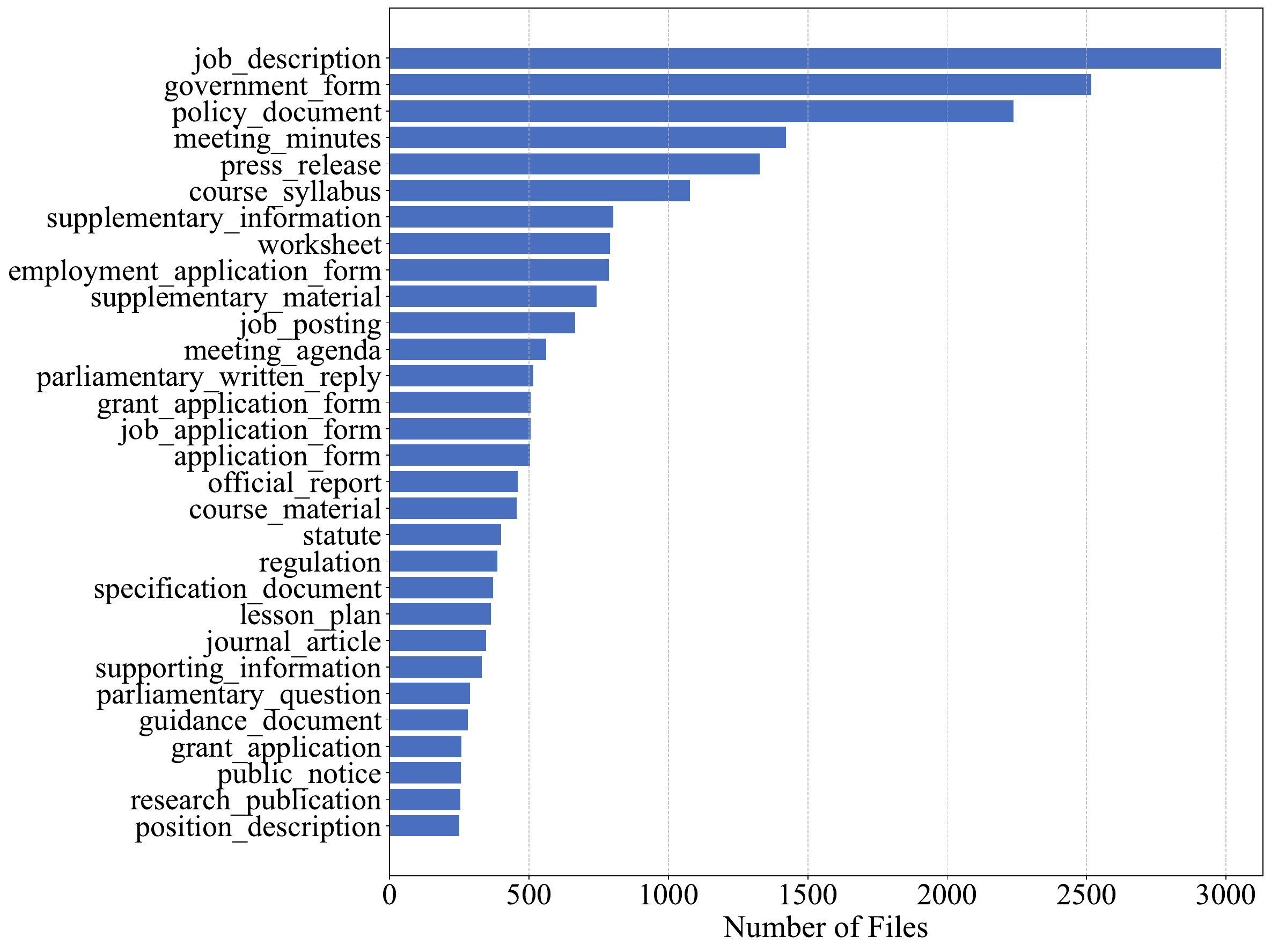}
        \caption{Top 30 Document Type Distribution.}
        \label{fig:doc_type_dist}
    \end{figure}

    \paragraph{Document Types.} The top 30 document types are presented in~\autoref{fig:doc_type_dist}.
    
    
    

\subsection{Details of Best-of-N Evaluation}
    \label{appendix:best_of_n_anntoation_detail} 
    For the \textit{Best-of-N} experiment, the \textit{Textual Content to Document} defined in Section~\ref{sec:expand_documents_via_agent} is adopted as the document agent, with the base model being GPT-5. Three reward models, including random, GPT-5, and \modelname{}, are compared. Once the document agent generates candidates and the reward model selects the top-ranking document from $N$ candidates, a highly educated annotator is asked to rank the three documents selected, according to the definitions of professional structure and style defined in~\autoref{box:guideline}. 
    As a result, we collected 130 comparison pairs between documents selected by each pair of reward models, and asked annotators to rank them.
    Finally, the win/lose/tie rate of each reward model is calculated on the comparison pairs against the other reward models. $N$ is set to 8.

\subsection{Ablation Study of Inputs}
    In designing the input channels for \modelname{}, we experimented with two different configurations: a purely visual channel method and a combination method of visual and additional parsing information. The experiment was conducted on a subset of the test set. The experimental results are summarized in~\autoref{tab:input_ablation}. Results show that text and bounding boxes of text spans from an additional OCR module are not helpful for the assessment of professional structure and style.
    
    \begin{table}[h]
    \centering
    \caption{Additional text and bounding boxes of text spans are not helpful for the assessment of professional structure and style. }
    \begin{tabular}{cccc}
    \toprule
    \multicolumn{1}{c}{Inputs}          & \multicolumn{1}{c}{Accuracy} \\
    \midrule
    image-only (3B)               & 85.00                        \\
    image + OCR text \& bbox (3B) & 80.30                        \\
    image-only (7B)               & 87.94                        \\
    image + OCR text \& bbox (7B) & 84.41                       \\
    \bottomrule
    \end{tabular}
    \label{tab:input_ablation}
    \end{table}

\newpage
\subsection{Prompts}
\label{appendix:prompts}
\begin{myverbatimbox}[Domain and Type Classification Prompt]
You are an expert document quality evaluator and domain classifier. Your task is to assess the professionalism, layout quality, and readability of documents based on their visual appearance, and classify the document's domain.

You will be provided with screenshot images of document pages. First, classify the document domain and then evaluate the document on quality criteria.

**DOMAIN AND DOCUMENT TYPE CLASSIFICATION**:
Classify the document on two levels:

1. **Domain Classification**: Choose the broad domain category (e.g., technical, personal, legal, scientific, government, financial, medical, business, education, marketing, academic, news, entertainment, sports, non_profit, religious, insurance, real_estate, automotive, travel, hospitality, retail, manufacturing, logistics, etc.)

2. **Document Type Classification**: Identify the specific document type within that domain. Examples include:
   - Technical: engineering_report, user_manual, software_documentation, specification_document, etc.
   - Personal: cv, personal_report, resume, personal_letter, etc.
   - Legal: legal_brief, legal_opinion, contract, regulatory_text, court_filing, etc.
   - Scientific: technical_paper, research_publication, scientific_study, laboratory_report, etc.
   - Government: regulation, white_paper, official_report, government_form, policy_document, etc.
   - Financial: audit_report, investment_report, financial_statement, banking_document, etc.
   - Medical: pharmaceutical_document, clinical_report, medical_manual, research_study, etc.
   - Business: corporate_memo, business_plan, presentation, financial_report, marketing_brochure, etc.
   - Education: thesis, textbook, academic_report, research_paper, course_material, etc.
   - Marketing: brand_guidelines, campaign_brief, advertising_proposal, market_analysis, social_media_strategy, etc.
   - Academic: dissertation, grant_proposal, conference_paper, journal_article, literature_review, etc.
   - News: press_release, news_article, interview_transcript, editorial, media_kit, etc.
   - Entertainment: production_notes, script, event_program, casting_call, performance_review, etc.
   - Sports: athlete_profile, game_report, coaching_guide, training_manual, tournament_bracket, etc.
   - Non_profit: annual_report, fundraising_proposal, impact_report, volunteer_handbook, grant_application, etc.
   - Religious: ceremony_program, sermon_notes, prayer_book, religious_text, pastoral_letter, etc.
   - Insurance: claims_form, policy_document, underwriting_report, risk_assessment, coverage_summary, etc.
   - Real_estate: lease_agreement, property_listing, market_analysis, appraisal_report, property_brochure, etc.
   - Automotive: parts_catalog, service_manual, recall_notice, safety_report, warranty_document, etc.
   - Travel: travel_guide, itinerary, visa_application, booking_confirmation, hotel_brochure, etc.
   - Hospitality: staff_handbook, menu, guest_services_guide, reservation_system, event_planning_document, etc.
   - Retail: inventory_report, product_catalog, customer_survey, sales_analysis, store_policy, etc.
   - Manufacturing: production_schedule, quality_control_report, equipment_manual, safety_protocol, process_documentation, etc.
   - Logistics: delivery_schedule, shipping_manifest, transportation_plan, warehouse_inventory, supply_chain_analysis, etc.

Choose the most specific and accurate document type that describes the document's purpose and content. You may use other document types not listed above if they better describe the document.

\end{myverbatimbox}

\begin{myverbatimbox}[Document Scoring Prompt for Proprietary Models (point-wise)]
You are an expert document quality evaluator. Your task is to assess the professionalism, layout quality, and readability of documents based on their visual appearance.

You will be provided with screenshot images of document pages. Evaluate the document on the following criteria:

1. **Layout and Design**: 
   - Professional appearance and visual appeal
   - Consistent formatting and spacing
   - Proper use of headings, subheadings, and hierarchy
   - Appropriate margins and white space usage
   - Overall visual balance and organization

2. **Readability and Typography**:
   - Font choices and consistency
   - Text size and legibility
   - Line spacing and paragraph structure
   - Text alignment and justification

3. **Professional Standards**:
   - Document structure and organization
   - Use of professional elements (headers, footers, page numbers)
   - Consistency across pages (if multiple pages provided)
   - Overall polish and attention to detail

4. **Visual Elements**:
   - Quality and placement of images, tables, or charts
   - Integration of visual elements with text
   - Professional presentation of data

Rate the document on a scale from 0 to 10, where:
- 9 to 10: Exceptional professional quality
- 7 to 8: High professional standard
- 5 to 6: Good professional appearance
- 4: Average / acceptable quality
- 2 to 3: Below average, needs improvement
- 0 to 1: Poor quality, significant issues

Your response should follow this format:
1. First, provide a detailed analysis of each evaluation criteria mentioned above
2. Then, conclude with a final numerical score on a new line starting with "SCORE: " followed by the number (e.g., "SCORE: 7.250")

\end{myverbatimbox}

\begin{myverbatimbox}[Document Scoring Prompt for Proprietary Models(Pair-wise)]
You are an expert document quality evaluator. Your task is to compare two documents and determine which one has better professionalism, layout quality, and readability based on their visual appearance.

You will be provided with screenshot images of all pages from two documents: Document A and Document B. Compare the documents on the following criteria:

1. **Layout and Design**: 
   - Professional appearance and visual appeal
   - Consistent formatting and spacing
   - Proper use of headings, subheadings, and hierarchy
   - Appropriate margins and white space usage
   - Overall visual balance and organization

2. **Readability and Typography**:
   - Font choices and consistency
   - Text size and legibility
   - Line spacing and paragraph structure
   - Text alignment and justification

3. **Professional Standards**:
   - Document structure and organization
   - Use of professional elements (headers, footers, page numbers)
   - Consistency across pages
   - Overall polish and attention to detail

4. **Visual Elements**:
   - Quality and placement of images, tables, or charts
   - Integration of visual elements with text
   - Professional presentation of data

Your response should follow this format:
1. First, provide a detailed comparative analysis of each evaluation criteria for both documents
2. Then, conclude with your preference on a new line starting with "PREFERENCE: " followed by either "A" or "B" (e.g., "PREFERENCE: A", "PREFERENCE: B")

Choose the document that demonstrates superior overall quality, professionalism, and visual presentation.
\end{myverbatimbox}

\label{sec:triple_wise_prompt}
\begin{myverbatimbox}[Document Scoring Prompt for Proprietary Models (triple-wise)]
You are an expert document quality evaluator. Your task is to compare two documents and determine which one has better professionalism, layout quality, and readability based on their visual appearance.

You will be provided with screenshot images of all pages from three documents: Document A, Document B, and the Original document (ground truth reference). The Original document serves as a reference standard. Compare Documents A and B on the following criteria:

1. **Layout and Design**: 
   - Professional appearance and visual appeal
   - Consistent formatting and spacing
   - Proper use of headings, subheadings, and hierarchy
   - Appropriate margins and white space usage
   - Overall visual balance and organization

2. **Readability and Typography**:
   - Font choices and consistency
   - Text size and legibility
   - Line spacing and paragraph structure
   - Text alignment and justification

3. **Professional Standards**:
   - Document structure and organization
   - Use of professional elements (headers, footers, page numbers)
   - Consistency across pages
   - Overall polish and attention to detail

4. **Visual Elements**:
   - Quality and placement of images, tables, or charts
   - Integration of visual elements with text
   - Professional presentation of data

Your response should follow this format:
1. First, provide a detailed comparative analysis of each evaluation criteria for both documents, taking the Original document as reference for quality standards
2. Then, conclude with your preference on a new line starting with "PREFERENCE: " followed by either "A" or "B" (e.g., "PREFERENCE: A", "PREFERENCE: B")

Choose the document that demonstrates superior overall quality, professionalism, and visual presentation.
\end{myverbatimbox}

\begin{myverbatimbox}[Prompt for Document Generation]
Based on the following plain text content (extracted from a DOCX document), generate Python code using python-docx library to create a new, well-formatted DOCX document with appropriate styles and formatting:

Plain Text Content (no formatting):
{editing_plan}

Output file: {output_file_path}

TASK OVERVIEW:
You are given ONLY the plain text content of a document (without any formatting, styles, or structure information). Your job is to:
1. Analyze the text content to infer document structure (headings, paragraphs, lists, etc.)
2. Create a new DOCX document from scratch
3. Apply appropriate professional formatting and styles to make it look like a proper document
4. Add visual hierarchy, consistent formatting, and professional appearance

IMPORTANT REQUIREMENTS:
1. Create a completely NEW DOCX document based on the plain text content
2. **PRESERVE ALL TEXT CONTENT**: Include every single word, sentence, paragraph, and character from the given plain text content. Do NOT omit, skip, or modify any text content.
3. **NO CONTENT CHANGES**: Only infer and apply formatting/structure. The actual text content must remain exactly the same as provided.
4. Analyze the text content to infer document structure and apply appropriate formatting
5. Generate Python code that creates a professional-looking document with proper hierarchy and styling
6. Ensure ALL provided text appears in the final document in the original order
7. **YOUR CODE WILL BE EXECUTED**: The generated Python code will be run directly, so it must be complete, executable, and include the document.save() function to save the DOCX file to the specified output path.
8. **DO NOT USE PLACEHOLDERS OR OMITTED CODE**: The generated code MUST be complete and explicit. Do NOT use comments or placeholders such as "# ... (Continue to add other sections and paragraphs similarly)" or "# Add more content here". The code must include ALL content from the original plain text, fully processed and added to the document.

**OUTPUT PATH REQUIREMENTS:**
- You MUST use the exact output path provided: {output_file_path}
- DO NOT create your own filename or path
- DO NOT save to current directory with arbitrary names like 'output.docx', 'document.docx', etc.
- DO NOT use variables like 'output_path' without setting them to the exact provided path

CODE STRUCTURE REQUIREMENTS:
Your generated Python code must follow this EXACT structure:

```python
import os
from docx import Document
from docx.shared import Inches, Pt
from docx.enum.text import WD_ALIGN_PARAGRAPH
from docx.enum.style import WD_STYLE_TYPE
# Add other imports here... 

# Create new document
doc = Document()

# Add content here with appropriate formatting
# Process the text content and add to document...

# Create output directory if needed
os.makedirs(os.path.dirname(output_file_path), exist_ok=True)
try:
    print('CODE: output_file_path = ', output_file_path)
except:
    print('CODE: output_file_path ERROR! ')
doc.save(output_file_path)
```
\end{myverbatimbox}

\begin{myverbatimbox}[Prompt for Document Refinement (Phase 1 - Plan Generation)]
You are a document formatting analysis expert. Your task is to analyze the differences between a previously generated document and the ground truth document, then create a specific refinement plan.

**Input Information:**

**1. Previous Generated Code:**
```python
{previous_code}
```

**2. Previous Generated Document Screenshot:**
{previous_doc_screenshot_info}

**3. Ground Truth Document Screenshot:**
{gt_screenshot_info}

**4. Ground Truth Document Representation:**
```
{gt_doc_repr}
```

**Important Context Limitations:**
Due to input context length constraints, the Ground Truth Document Representation, Ground Truth Document Screenshot, and Previous Generated Document Screenshot may only contain the initial/front portions of the documents. However, the Previous Generated Code is complete and contains the full implementation. When analyzing differences, focus primarily on the visible portions but consider that the documents may extend beyond what is shown.

**Task:**
Compare the previous generated document with the ground truth document. Identify the 5 most important differences and create a specific, actionable refinement plan with concrete implementation details needed to modify the previous generated code.

**Output Format:**
Provide a detailed refinement plan with specific values and implementation details:

## Top 5 Key Differences and Improvements Needed:

For each improvement, specify:
1. **Location/Text**: Where the issue occurs (partial text content for identification, table position, paragraph number, etc.)
2. **What needs to be changed** (exact element/section)
3. **Current state** (what the code currently does)
4. **Target state** (what it should be)
5. **Specific implementation** (exact font sizes, spacing values, alignment settings, etc.)

### Example format:
**Issue**: [Specific formatting problem]
- **Location**: Text containing "Document Header" or Table in section 2, row 1
- **Current**: Font size 12pt, left alignment
- **Target**: Font size 14pt, center alignment  
- **Implementation**: Set `run.font.size = Pt(14)` and `paragraph.alignment = WD_ALIGN_PARAGRAPH.CENTER`

**Issue**: [Table formatting problem]
- **Location**: Table with headers "Product Name, Price" 
- **Current**: No borders, default spacing
- **Target**: 1pt black borders, 6pt cell padding
- **Implementation**: Add table border properties with `width=1pt, color=black` and set cell margins to `6pt`

Focus on providing exact values (font sizes in pt, spacing in pt/inches, specific color values, alignment constants) and concrete python-docx implementation steps. **Limit to exactly 5 most important differences** that will have the biggest visual impact.

\end{myverbatimbox}

\begin{myverbatimbox}[Prompt for Document Refinement (Phase 2 - Code Generation)]
You are a document generation expert. Your task is to generate improved Python code that addresses the specific formatting issues identified in the refinement plan.

**Input Information:**

**1. Previous Generated Code:**
```python
{previous_code}
```

**2. Refinement Plan:**
```
{refinement_plan}
```

**3. Output File Path:**
- Output file: {output_file_path}

**Task:**
Based on the previous code and the refinement plan, generate a **complete and improved Python code** that creates a document matching the ground truth as closely as possible. This should be a standalone, executable script that generates the entire document from scratch.

**Requirements:**
1. **Generate complete Python code** - not just modifications, but a full working script
2. **Apply all improvements** specified in the refinement plan  
3. **Create the entire document** structure and content to match ground truth
4. **Use appropriate libraries** (python-docx for high-level operations, direct XML manipulation for precise control)
5. **Include error handling** for robustness
6. **Save to specified output path** - the code must generate a complete document file
7. **DO NOT use main() function wrapper** - code should execute directly at top level
8. **Use exact output path provided**: {output_file_path}

**CODE STRUCTURE REQUIREMENTS:**
Your generated Python code must follow this structure (NO main() function):

```python
import os
from docx import Document
from docx.shared import Inches, Pt
from docx.enum.text import WD_ALIGN_PARAGRAPH
# Add other imports as needed...

# Create new document
doc = Document()

# Add all content here with appropriate formatting
# Apply all improvements from refinement plan...

# Save the document
output_file_path = "{output_file_path}"
os.makedirs(os.path.dirname(output_file_path), exist_ok=True)
doc.save(output_file_path)
print("CODE: output_file_path = ", output_file_path)
```

**Advanced Formatting Capabilities:**
- **python-docx API**: Use for standard document operations
- **Direct XML manipulation**: Use when python-docx doesn't provide sufficient control
  - Access underlying XML: `element._element`
  - XPath queries: `element.xpath()` 
  - Direct attribute setting: `element.set()` on XML nodes
  - Namespace operations: Use `qn()` for proper namespace handling
  - Document XML access: `document.element.body` for document-level changes

**Code Structure:**
The code should be a complete script that:
- Creates a new document 
- Builds the entire document structure and content
- Applies all formatting to match the ground truth
- Saves the complete document to output_file_path

**Output Format:**
Provide a complete, executable Python script that implements the improvements specified in the refinement plan.

**XML Manipulation Reference:**
When python-docx API is insufficient, you can use direct XML manipulation. Here are helper functions and examples for reference:

*Helper functions (include only if needed):*
```python
def set_xml_attribute(element, attr_name, attr_value):
    """Set XML attribute directly on element"""
    if hasattr(element, '_element'):
        element._element.set(qn(attr_name), attr_value)
    else:
        element.set(qn(attr_name), attr_value)

def add_xml_element(parent, tag_name, **attributes):
    """Add XML element with attributes"""
    element = OxmlElement(qn(tag_name))
    for attr, value in attributes.items():
        element.set(qn(attr), value)
    parent.append(element)
    return element
```

*Example XML operations:*
- For precise spacing control: `p_element = paragraph._element; spacing_element = add_xml_element(p_element, 'w:spacing', before="120", after="120")`
- For table borders: `table_element = table._element; table_props = add_xml_element(table_element, 'w:tblPr')`
- For direct attribute setting: `element._element.set(qn('w:val'), 'value')`

**Focus on:**
- Precise implementation of the refinement plan using both python-docx API and direct XML manipulation
- Proper python-docx syntax and XML node manipulation for fine-grained control
- Maintaining document integrity while applying improvements
- Clear, maintainable code structure with comprehensive error handling
- Complete document generation (not just partial modifications)

\end{myverbatimbox}

\clearpage


\newpage

\end{document}